\documentclass[10pt,twocolumn,letterpaper]{article}

\usepackage[pagenumbers]{cvpr}              
\definecolor{cvprblue}{rgb}{0.21,0.49,0.74}
\usepackage{color,xcolor,colortbl}
\usepackage[pagebackref,breaklinks,colorlinks,allcolors=cvprblue]{hyperref}

\newcommand\B[1]{\textcolor{blue}{#1}}
\newcommand\R[1]{\textcolor{red}{#1}}
\DeclareUnicodeCharacter{FF1A}{:}
\usepackage{amssymb}
\usepackage{caption}
\usepackage{adjustbox}
\usepackage{graphicx}
\usepackage{tabularx}
\usepackage[utf8]{inputenc} 
\usepackage[T1]{fontenc}    
\usepackage{url}            
\usepackage{booktabs}       
\usepackage{amsfonts}       
\usepackage{nicefrac}       
\usepackage{microtype}      
\usepackage{xcolor}         
\usepackage{amsmath}
\usepackage{multirow}
\usepackage{natbib}
\usepackage{wrapfig}
\usepackage{float}
\setcitestyle{numbers,square}
\def\paperID{7850} 
\def\confName{CVPR}
\def\confYear{2026}
\title{QuantFace: Efficient Quantization for Face Restoration}

\author{
  \textbf{Jiatong Li$^{1}$\thanks{Equal contribution.},\enspace Libo Zhu$^{1}$\footnotemark[1],\enspace Haotong Qin$^{2}$,\enspace Jingkai Wang$^{1}$,\enspace Linghe Kong$^{1\dagger}$,}\\ \textbf{Guihai Chen$^{1}$,\enspace Yulun Zhang$^{1}$\thanks{Corresponding authors: Yulun Zhang, yulun100@gmail.com; Linghe Kong, linghe.kong@sjtu.edu.cn},\enspace Xiaokang Yang$^{1}$} \\
  \textsuperscript{1}Shanghai Jiao Tong University,\\
  \textsuperscript{2}ETH Z\"{u}rich
}

\begin{document}
\maketitle
\begin{abstract}
Diffusion models have been achieving remarkable performance in face restoration. However, the heavy computations hamper the widespread adoption of these models. In this work, we propose QuantFace, a novel low-bit quantization framework for face restoration models, where the full-precision (\ie, 32-bit) weights and activations are quantized to 4$\sim$6-bit. We first analyze the data distribution within activations and find that it is highly variant. To preserve the original data information, we employ rotation-scaling channel balancing. Furthermore, we propose Quantization-Distillation Low-Rank Adaptation (QD-LoRA), which jointly optimizes for quantization and distillation performance. Finally, we propose an adaptive bit-width allocation strategy. We formulate such a strategy as an integer programming problem that combines quantization error and perceptual metrics to find a satisfactory resource allocation. Extensive experiments on the synthetic and real-world datasets demonstrate the effectiveness of QuantFace under 6-bit and 4-bit. QuantFace achieves significant advantages over recent leading low-bit quantization methods for face restoration.
\end{abstract}    
\section{Introduction}
Face restoration seeks to recover high-quality (HQ) facial images from degraded low-quality (LQ) inputs. The LQ inputs are generated from complex degradation processes, such as blurring, noise, downsampling, and JPEG compression. With the rapid development of deep generative models, generative adversarial networks (GANs)~\cite{goodfellow2014generative} and diffusion-based methods~\cite{ddpm,ddim,song2020score} have attracted increasing attention. Several approaches~\cite{kawar2022ddrm,wang2022ddnm,fei2022gdp,yang2023pgdiff} have already demonstrated impressive capability in restoring HQ human faces. Nevertheless, GANs typically suffer from training instability due to adversarial learning, while multi-step diffusion models demand extensive computational resources.

To accelerate inference, recent studies have proposed one-step diffusion (OSD)~\cite{dmd,song2023consistency,li2024distillation} schemes that reduce the denoising step to one. At the same time, these models largely preserve the performance of the original multi-step diffusion models. Thanks to these advanced techniques, one-step diffusion face restoration (OSDFR) models~\cite{osdface} have become feasible. However, despite this progress, deploying such models on resource-constrained platforms (\eg, smartphones or other mobile devices) remains challenging, because they still require prohibitive computational resources. This limitation hinders the widespread adoption of diffusion-based face restoration methods in real-world scenarios.

\begin{figure}
\scriptsize
\vspace{-3.5mm}
\centering

\begin{adjustbox}{valign=t}
{\setlength{\tabcolsep}{1pt}
\begin{tabular}{@{}ccccc@{}}
\includegraphics[width=0.09\textwidth]{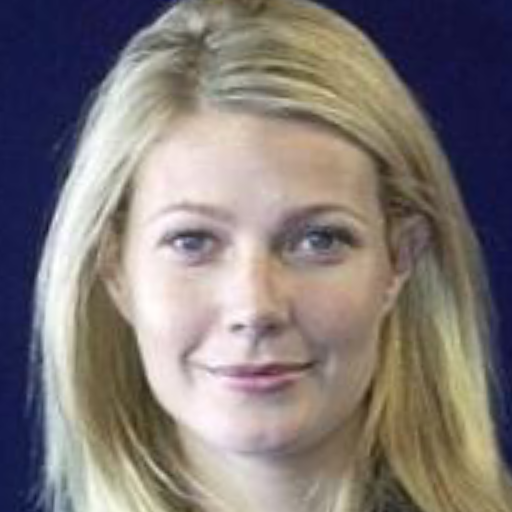} &
\includegraphics[width=0.09\textwidth]{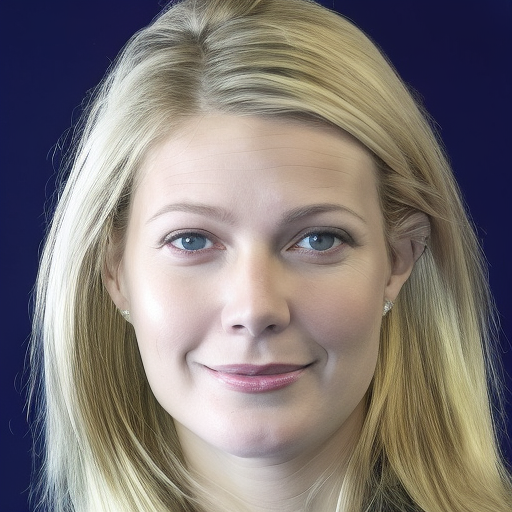} &
\includegraphics[width=0.09\textwidth]{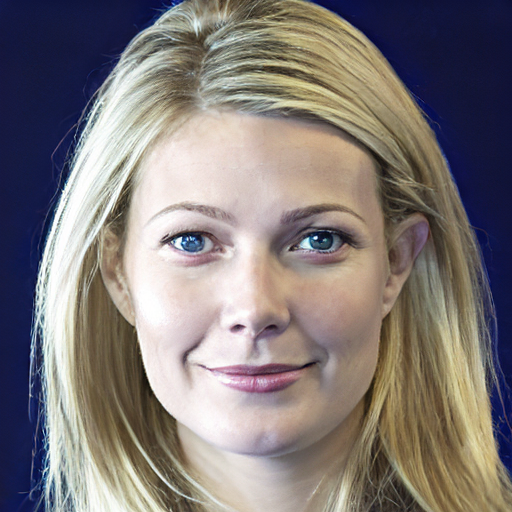} &
\includegraphics[width=0.09\textwidth]{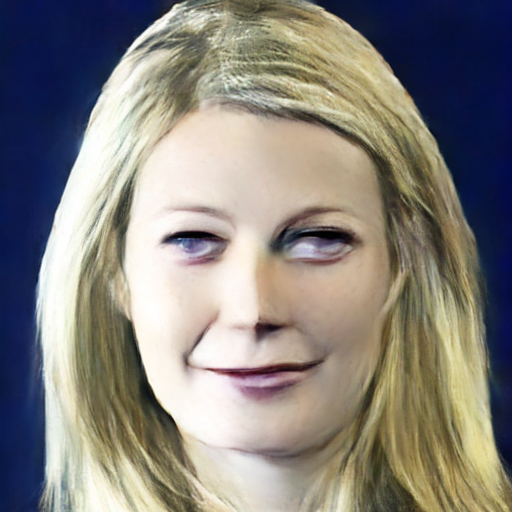} &
\includegraphics[width=0.09\textwidth]{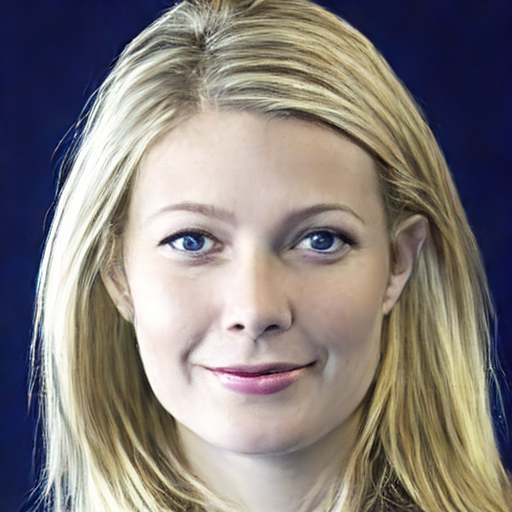} \\
LQ &
DiffBIR~\cite{lin2024diffbir} &
OSDFace~\cite{osdface} &
SVDQuant~\cite{li2025svdquant} &
QuantFace (ours) \\
\# Step / Bits &
50 / 32-bit &
1 / 32-bit &
1 / 4-bit &
1 / 4-bit \
\end{tabular}}
\end{adjustbox}

\vspace{-3mm}
\caption{
Visual comparison between the multi-step diffusion and one-step diffusion face restoration models in full-precision, recent quantization methods in 4-bit, and our QuantFace in 4-bit. Our method achieves an 84.85\% parameter compression and an 82.91\% speedup compared with the full-precision OSDFace~\cite{osdface}.
}
\vspace{-3.5mm}
\label{fig:poster}
\end{figure}

Model quantization~\cite{white} offers an effective avenue to further accelerate OSDFR models. By mapping both weights and activations from full-precision (FP) to low-bit precision, quantization significantly reduces memory footprint and computational overhead. Nevertheless, the performance gap between the quantized and the FP version is inevitable, especially under the aggressive 4-bit setting. Minimizing this gap is therefore vital for the successful implementation of quantized OSDFR models.

Although current low-bit quantization strategies have achieved promising results for multi-step diffusion models and for text-to-image generation~\cite{qdiffusion,PTQD,PTQ4DM,li2025svdquant,sui2024bitsfusion,He2023EfficientDM}, significant performance drops occur when we apply these methods to OSDFR models.  There are three main challenges in low-bit quantization for OSDFR. \textbf{First}, the distribution of dynamic activation values is vital for high-frequency facial details but it is poorly preserved by static quantizers. \textbf{Second}, current calibration schemes from FP to quantized models are suboptimal. \textbf{Third}, the limited computational resources are allocated improperly.

In this work, we propose QuantFace, a unified and effective low-bit quantization framework for OSDFR models. We select OSDFace~\cite{osdface} as our quantization backbone due to its excellent performance. To address the aforementioned challenges, we propose tailored solutions for each. \textbf{First}, we conduct a detailed analysis of the activation distributions in OSDFace and find that different channels of the activation values exhibit high variance. Previous scaling-based methods~\cite{smoothquant,li2025svdquant} can remove the outliers from activations to weights and thus decrease quantization error. 
However, these methods are ineffective in preserving the original activation distribution after per-tensor quantization. We find that the distribution information is vital for basic facial structure generation. 
To solve the problem, we propose rotation-scaling channel balancing. We consider rotation and scaling simultaneously to maintain a uniform activation distribution. \textbf{Second}, to promote alignment between the FP model and the quantized model, we propose Quantization-Distillation Low-Rank Adaptation (QD-LoRA). QD-LoRA effectively compensates for quantization error while facilitating distillation between the FP model and the quantized model. \textbf{Third}, we propose adaptive bit-width allocation. We analyze the layer-wise sensitivity to quantization and configure the bottlenecks. We design an integer programming-based algorithm for mixed-precision activation configuration. This algorithm incorporates perceptual metrics and numeric reconstruction error as the optimization target and promotes reasonable resource allocation.

We have conducted extensive evaluations on both synthetic and real-world datasets. As shown in Fig.~\ref{fig:poster}, even under the aggressive 4-bit setting, our model achieves visual quality comparable to the multi-step diffusion model DiffBIR~\cite{lin2024diffbir}. Furthermore, Tab.~\ref{tab:complexity} demonstrates that at 4-bit precision, we attain parameter and computational cost compression ratios of 84.85\% and 82.91\%, respectively, compared with OSDFace~\cite{osdface}. These findings validate the efficacy of our proposed approaches for quantizing OSDFR models.

In summary, our contributions are as follows:
\begin{itemize}
    \vspace{-0.5mm}
    \item We conduct a detailed analysis of the activation distribution of the OSDFace and identify that quantization of activation values is crucial for preserving facial features. We configure that the inadequate processing of highly variant activations is the key reason for the performance degradation in quantization.
    \item We propose QuantFace, a unified quantization framework for OSDFR models. It consists of rotation-scaling channel balancing,  QD-LoRA, and adaptive bit-width allocation.
    \item Our rotation-scaling channel balancing strategy effectively preserves the original activation information, while QD-LoRA enhances the alignment between the quantized model and FP model. Meanwhile, adaptive bit-width allocation facilitates reasonable usage of resources.
    \item Extensive experiments on the synthetic and real-world datasets demonstrate the superior performance of our method. Quantitative and qualitative results show that our method significantly outperforms existing quantization approaches. Under the extreme 4-bit setting, our model is still capable of generating high-quality face images.
\end{itemize}

\begin{figure}
\centering
\small
\resizebox{\linewidth}{!}{
\includegraphics[width=0.15\textwidth]{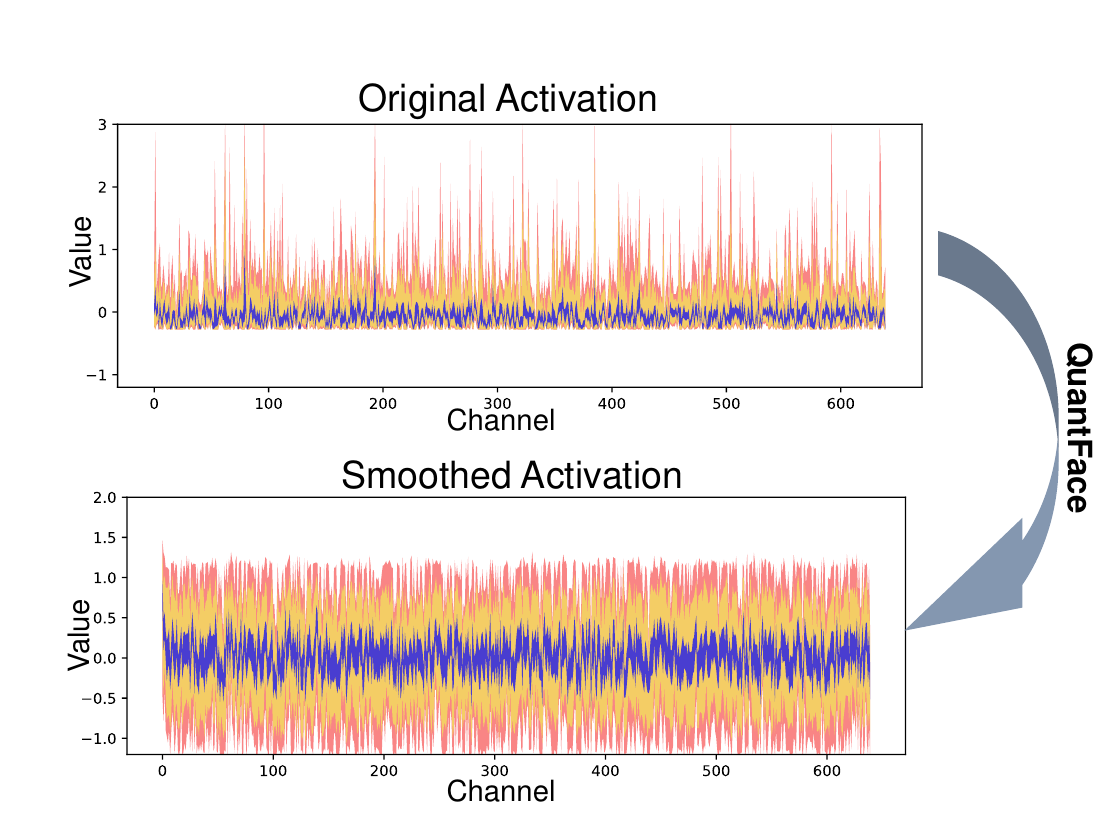} 
}
\caption{The original activation has high variance. Our QuantFace can smooth activation distribution and reduce quantization error.}
\vspace{-3.5mm}
\label{fig:QuantFace}
\end{figure}
\begin{figure*}[!h]
    \centering
    \includegraphics[width=\textwidth]{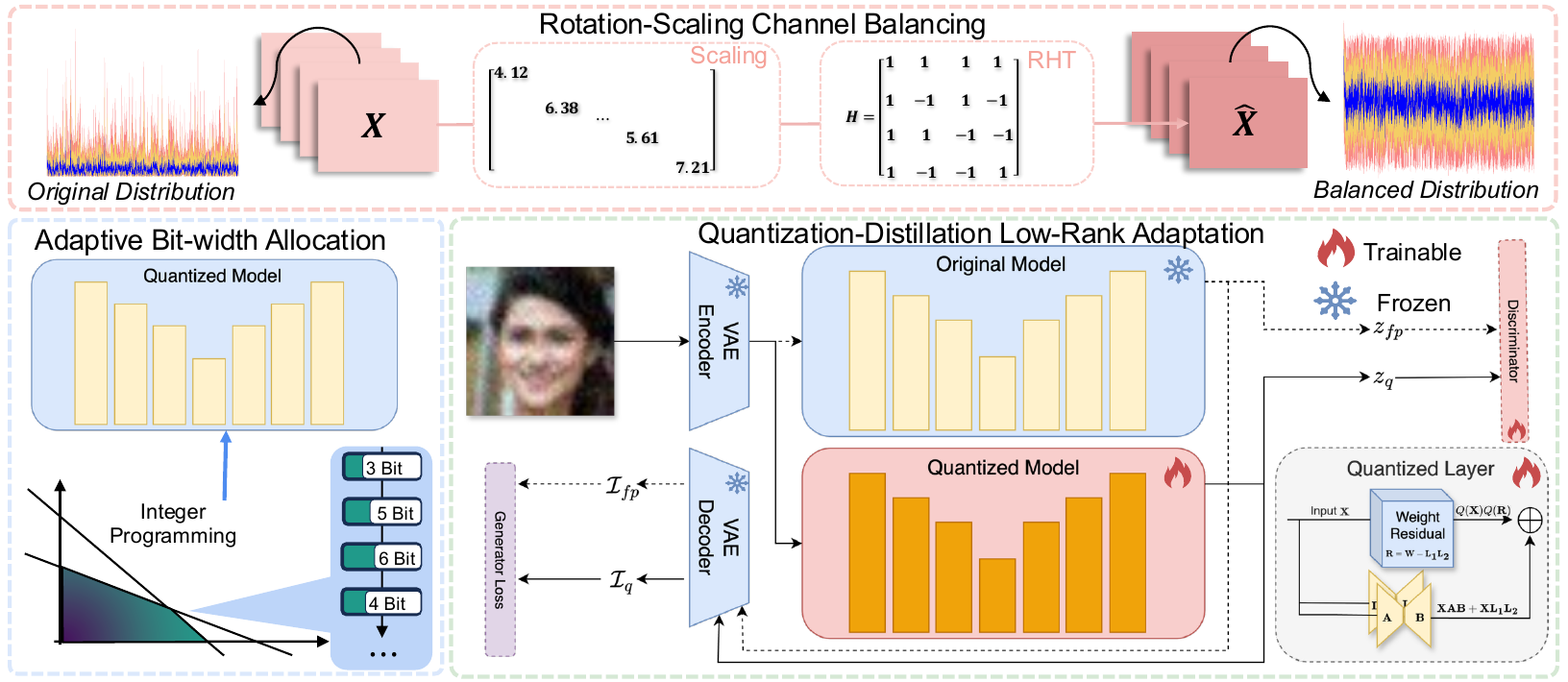}
    \vspace{-6mm}
    \caption{Overview of our QuantFace. \textbf{First}, under the 4-bit precision setting, we use the quantization errors and perceptual importance weights as the objective for integer programming, and allocate appropriate precision to activation of each layer. \textbf{Second}, before training, we integrate the scaling factor and rotation matrix into the weights, and only apply an online RHT in the convolution layers. \textbf{Third}, we align the quantized model with the FP model on the calibration dataset by optimizing the dual-branch low-rank matrices we design.}
    \label{fig:total}
    \vspace{-5mm}
\end{figure*}
\section{Related Works}
\subsection{Face Restoration}
\vspace{-2mm}
Face restoration seeks to reconstruct high-fidelity facial images from degraded low-quality inputs affected by diverse and complex distortions. A central challenge in this domain is how to incorporate facial priors efficiently and effectively. While classical methods have long leveraged statistical priors, recent advances in deep generative techniques (such as GANs~\cite{goodfellow2014generative} and diffusion models~\cite{ddpm,ddim,rombach2022highresolutionimagesynthesislatent}) have driven growing interest in generative priors.

In particular, diffusion-based priors~\cite{kawar2022ddrm,wang2022ddnm,fei2022gdp,yang2023pgdiff} have attracted considerable attention. These methods generate HQ images by gradually removing noise from LQ inputs. Concurrently, vector-quantized (VQ) prior approaches~\cite{vqfr,zhou2022codeformer,wang2023restoreformer++,daefr} are promising. They map LQ features to codebook entries and then decode the HQ images from the codebook.

The combination of VQ priors and generative diffusion priors has become very popular recently. However, while these hybrid methods can generate excellent human face images, the computational demands become severe. Consequently, developing effective compression techniques for these models has become a major focus of current research.

\subsection{Model Quantization}
\vspace{-2mm}
Model quantization improves computational efficiency by reducing the precision of model parameters while maintaining performance. Quantization methods are broadly classified into two categories based on whether weight retraining is involved: post-training quantization (PTQ)~\cite{ZeroQuant,brecq} and quantization-aware training (QAT)~\cite{Bhalgat2020LSQ+,efficientqat,yu2024improving}. Furthermore, quantization has been demonstrated as an effective technique for compressing large language models (LLMs)~\cite{liu2023llm,bondarenko2024low,li2023loftq,smoothquant} for deployment on edge devices. With the rapid advancements in diffusion models (DMs), there has been a growing focus on enhancing their efficiency through quantization techniques including PTQ methods such as PTQ4DM~\cite{PTQ4DM}, Q-Diffusion~\cite{li2023q}, and PTQD~\cite{PTQD}, as well as QAT methods like Q-DM~\cite{qdm}. Recently, EfficientDM~\cite{He2023EfficientDM} introduced a low-rank quantization fine-tuning strategy, while SVDQuant~\cite{li2025svdquant} employs 16-bit parallel low-rank branches to preserve performance. PassionSR~\cite{zhu2024passionsr} presents an innovative quantization strategy, achieving 8-bit and 6-bit quantization for one-step diffusion-based super-resolution (OSDSR) models. However, existing works on quantizing image restoration models to 4-bit remain limited.

\begin{figure*}[t]
    \centering
    \begin{minipage}{0.5\textwidth}
        \captionsetup{font=footnotesize}
        \centering
        \caption*{\textbf{Original Activation}}
        \vspace{-2.8mm}
        \includegraphics[width=\textwidth]{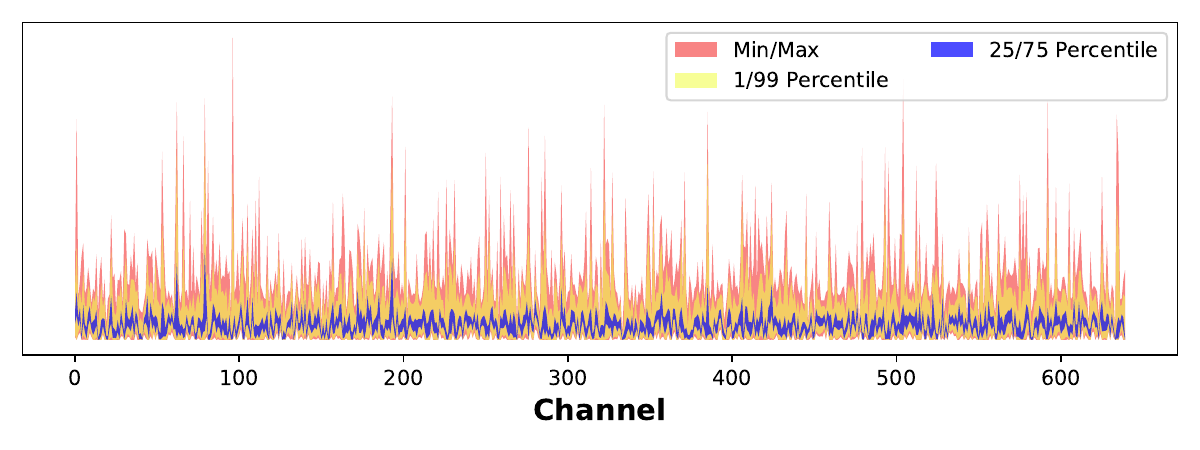}
    \end{minipage}%
    \begin{minipage}{0.5\textwidth}
        \captionsetup{font=footnotesize}
        \centering
        \caption*{\textbf{Channel-wise Scaling}}
        \vspace{-2.8mm}
        \includegraphics[width=\textwidth]{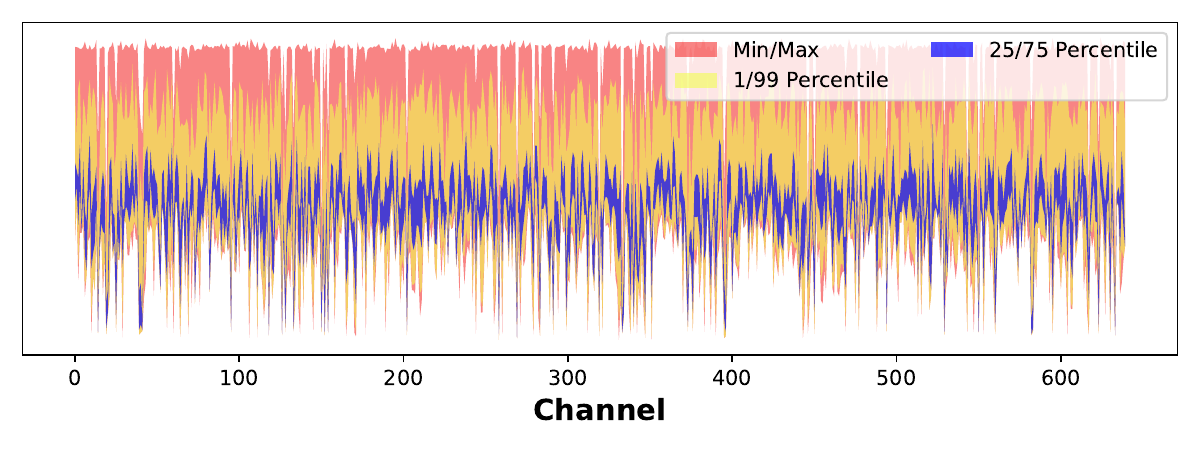}
    \end{minipage} \\
    \begin{minipage}{0.5\textwidth}
        \captionsetup{font=footnotesize}
        \caption*{\textbf{Randomized Hadamard Transformation}}
        \vspace{-2.8mm}
        \centering
        \includegraphics[width=\textwidth]{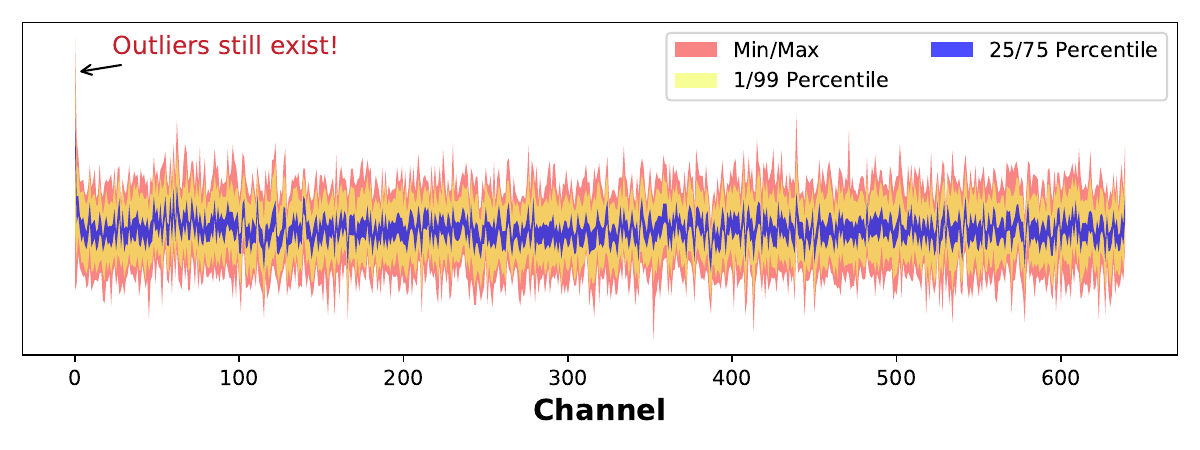}
    \end{minipage}%
    \begin{minipage}{0.5\textwidth}
        \captionsetup{font=footnotesize}
        \caption*{\textbf{Rotation-Scaling Channel Balancing}}
        \vspace{-2.8mm}
        \centering
        \includegraphics[width=\textwidth]{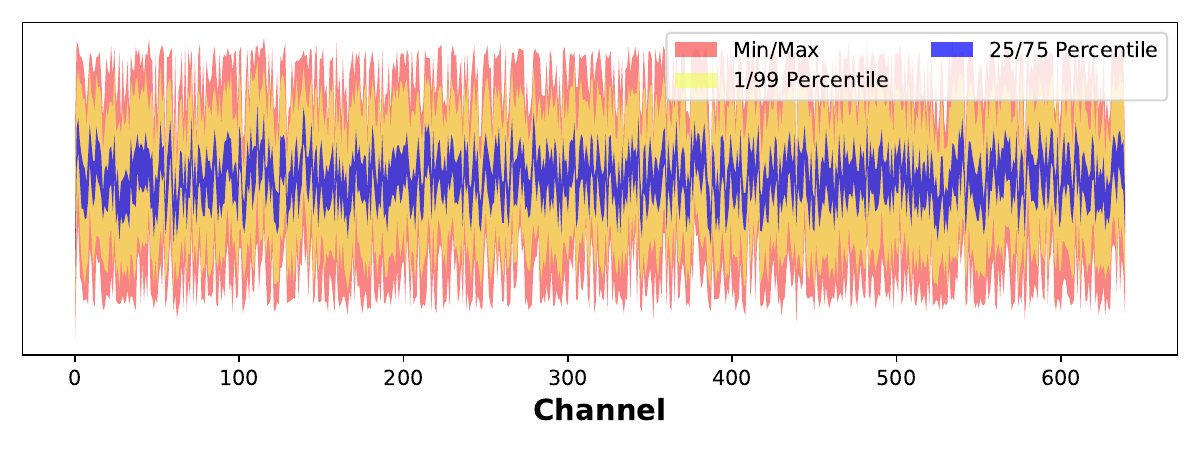}
    \end{minipage}
    \vspace{-3.8mm}
    \caption{A sample data distribution along input channels of unet$.$down\_blocks$.$1$.$resnets$.$1$.$conv2 activations. They are processed with different balancing techniques. Directly employing the Scaling or RHT will not result in satisfactory distribution for quantization.}
    \label{fig:four_images}
    \vspace{-3mm}
\end{figure*}
\section{Methods}
\subsection{Preliminaries}
\vspace{-2mm}
\paragraph{Diffusion Model.}
Diffusion models~\cite{rombach2022highresolutionimagesynthesislatent} employ a two‐stage procedure. First, a complex data distribution is gradually corrupted by successive injections of Gaussian noise. Second, a neural network learns the inverse of this noising process to restore the original samples. In the forward diffusion, the original sample $\mathbf{x}_0$ is progressively transformed into pure Gaussian noise $\mathbf{x}_T$ by injecting random noise over $T$ steps, formally defined as~\cite{ddim}:
\begin{equation}
\mathbf{x}_t = \sqrt{\bar{\alpha}_t} \, \mathbf{x}_0 + \sqrt{1 - \bar{\alpha}_t}\, \epsilon, \quad \epsilon\sim\mathcal{N}(0,1),
\label{equ: diffusion_base}
\end{equation}
where $ \bar{\alpha}_t$ is a scheduling parameter, and $t$ is the time step.

After $T$ iterations, the 
data pattern converges to a standard Gaussian distribution: $ \mathbf{x}_T \sim \mathcal{N}(0, \mathbf{I}) $. Reversed diffusion process reconstructs the original distribution by predicting the noise $\epsilon_\theta(\mathbf{x}_t,t)$ formulated as follows~\cite{ddim}:
\begin{equation}
\begin{aligned}
\mathbf{x}_{t-1}=\sqrt{\bar\alpha_{t-1}}(\frac{\mathbf{x}_t-\sqrt{1-\bar\alpha_t}\epsilon_\theta(\mathbf{x}_t,t)}{\sqrt{\bar\alpha_t}}) \\
+\sqrt{1-\bar\alpha_{t-1}-\sigma_t^2}\cdot\epsilon_\theta(\mathbf{x}_t,t)+\sigma_t\epsilon_t,
\end{aligned}
\label{equ: diffusion_base2}
\end{equation}
where $\epsilon_\theta(\mathbf{x}_t,t)$ represents the output of the noise prediction network, $\sigma^2_t$ denotes the noise variance at step $t$, and $\epsilon_t$ is random noise irrelevant to $\mathbf{x}_t$. By gradually predicting the random noise at timestep $t$, the model progressively optimizes 
 $\mathbf{x}_t$ to $\mathbf{x}_0$, and ultimately generates high-quality samples. Our one-step estimation~\cite{dmd} can be formulated as:
\begin{equation}
\mathbf{x}_0\approx\frac{\mathbf{x}_t-\sqrt{1-\bar\alpha_t}\epsilon_\theta(\mathbf{x}_t,t)}{\sqrt{\bar\alpha_t}}
\label{eq:onestep}
\end{equation}

\vspace{-3mm}
\paragraph{Model Quantization.}
Model quantization~\cite{jacob2017quantizationtrainingneuralnetworks} is an effective alternative to reduce memory and computational costs by converting floating-point numbers to fixed-point numbers. This process utilizes approximation with integer $x_{\text{int}}$ and quantization parameters (scaling factor $s$, zero point $z$). The fake quantization~\cite{white} process can be defined as:
\begin{equation}
x_{\text{int}} = \text{clamp}\left(\left \lfloor\frac{ \mathbf{x}}{{s}} \right \rceil-z , l, u \right), \hat{x} = s \cdot x_{\text{int}} + z,
\label{equ: q_base}
\end{equation}
where $x_{\text{int}}$ refers to the quantized integer and $\hat{x}$ is the simulated quantized float-point number. $\left \lfloor \cdot\right \rceil$ is the round-to-nearest operator, and $ \text{clamp}(\cdot, l, u) $ ensures values remain within the range $[l, u]$.

To address the non-differentiability of the rounding operation, we employ the straight-through estimator (STE~\cite{liu2022nonuniform}) to approximate the gradients during backpropagation:
\begin{equation}
\frac{\partial Q(x)}{\partial x} \approx 
\begin{cases} 
1 & \text{if } x \in [l, u], \\
0 & \text{otherwise}.
\end{cases}
\label{equ:STE}
\end{equation}
\vspace{-2mm}
\begin{figure*}[t]
    \centering
    \includegraphics[width=\textwidth]{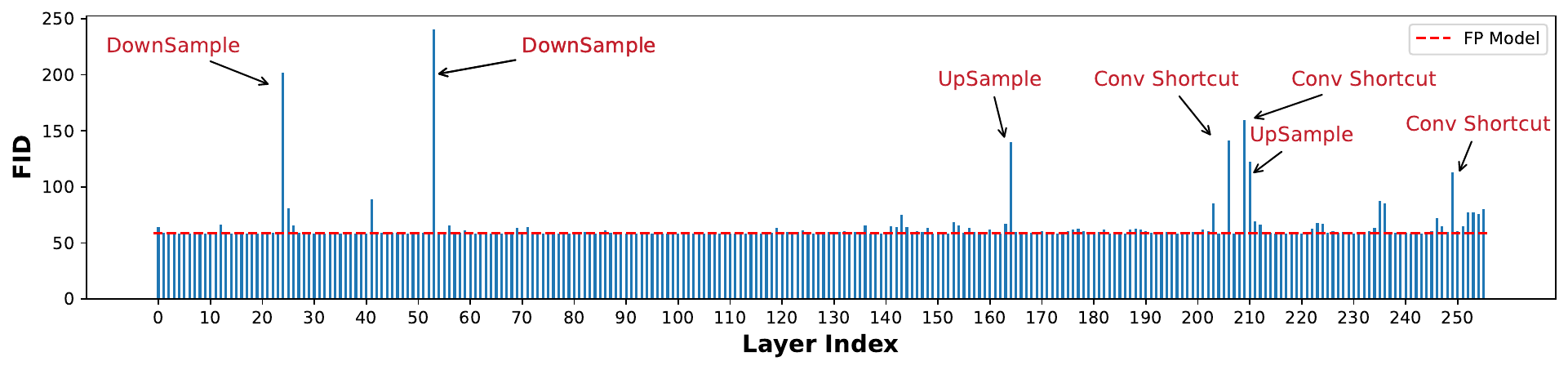}  
    \vspace{-6mm}
    \caption{The change in FID on FFHQ when quantizing the activation of each layer individually at 4-bit precision. Different layers exhibit varying levels of sensitivity to quantization. Downsampling, upsampling, and residual connections are bottlenecks for quantization.
}
\vspace{-5mm}
\label{fig:fid_sensitivity}
\end{figure*}

\subsection{Rotation-Scaling Channel Balancing}
\vspace{-2mm}
In any quantization method, appropriately managing the relationship between outliers and non‑outliers is of critical importance. Existing channel-scaling methods can shift the quantization difficulty from activations to weights~\cite{li2025svdquant, smoothquant}. After scaling, activations exhibit reduced magnitude and fewer outliers, resulting in lower quantization error. However, when it comes to convolution neural networks like UNet~\cite{unet}, we typically employ per‑tensor quantization for activations, which means a single set of quantization parameters is shared across all channels of a tensor. Although the aforementioned scaling method suppresses extreme outliers, inter-channel imbalance remains. When the bit‑width is extremely low (\eg, 4-bit), the performance of the quantized model drops significantly.

To solve the problem, incoherent processing via Randomized Hadamard Transform (RHT) has recently emerged as a promising technique~\cite{quipsharp}, and it is a remarkable quantization method for LLMs~\cite{quarot, spinquant} and ViTs~\cite{vidit}. This approach requires no parameter tuning and can effectively generate a uniform data distribution across channels, thereby maximally preserving information after quantization. However, this method cannot adequately smooth extreme outliers.

To overcome these limitations, we propose rotation-scaling channel balancing. We find that simultaneously considering scaling and rotation techniques can reduce quantization error effectively. And this method maximally preserves the distribution information of the original data. Figure~\ref{fig:four_images} illustrates the efficacy after applying the rotation-scaling channel balancing.

To the best of our knowledge, prior quantization works have primarily focused on Transformer architectures. We are the first to design a scaling and rotation channel balancing method for convolution neural networks like UNet~\cite{unet}. Through our analysis, we show that convolution can be transformed to matrix multiplication to some degree. The original 2D-convolution can be written as:
\begin{equation}
\begin{aligned}
y_{f,i,j}&=\displaystyle\sum_{u=1}^{K_w}\sum_{v=1}^{K_h}\sum_{c=1}^{C}k_{f,c,u,v}x_{c,i+u-1,j+v-1} \\&=\displaystyle\sum_{u=1}^{K_w}\sum_{v=1}^{K_h}\textbf{k}_{f,u,v}^{T}\textbf{x}_{i+u-1,j+v-1},
\label{eq:convlution}
\end{aligned}
\end{equation}
where \(C, K_w,k_h\) represents the number of input channels, kernel width, and kernel height. \(\textbf{k}_{f,u,v},\textbf{x}_{h+u-1,w+v-1}\) are tensors with 
$C$ elements. For simplicity, let:  
\begin{align}
    \textbf{k}_{f,u,v} \triangleq  \textbf{k}\in \mathbb{R}^{C \times 1},\textbf{x}_{h+u-1,w+v-1} \triangleq  \textbf{x}\in \mathbb{R}^{C\times 1},
\end{align}
so we have：
\begin{equation}
\begin{aligned}
   \textbf{k}^T\textbf{x}&=(\textbf{k}^T\odot s)(\frac{1}{s}\odot \textbf{x})=\textbf{k}^{T}{\mathbf{H}}^T\mathbf{H}\mathbf{x}, \\s&=\frac{\underset{(h, w)}{\max}(|\mathbf{X}|)^\alpha}{\underset{(C,K_w,K_h)}{\max}(|\mathbf{K}|)^{1-\alpha}},
\end{aligned}
\end{equation}
where $\mathbf{K}\in \mathbb{R}^{F\times C\times K_w\times K_h}, \mathbf{X}\in \mathbb{R}^{C\times h\times w}$ is convolution kernel and input tensor of the layer. $\mathbf{H}$ is a Hadamard matrix. We find that for convolution layers in OSDFace~\cite{osdface}, the corresponding Hadamard matrices exist. According to~\cite{quipsharp}, only a minimal amount of memory is required to store the corresponding random $\pm 1$ vector of each matrix.

To minimize the additional computational overhead, inspired by the offline transformation strategies from~\cite{quarot, Xu2025MambaQuantQT}, we integrate most of the scaling factors and rotation matrices into model weights, thereby minimizing the quantization overhead. During inference, we only perform online RHT for input activations of convolution layers.

\subsection{Quantization-Distillation LoRA}
\vspace{-2mm}
Under extremely low bit-width configurations, face restoration models struggle to generate fine facial detail features. Incorporating a high-precision branch to mitigate quantization errors is a widely adopted strategy~\cite{li2025svdquant,He2023EfficientDM}.
SVDQuant~\cite{li2025svdquant} shows that initializing these branches via singular value decomposition (SVD)~\cite{svd} has proven to be particularly effective. Inspired by SVDQuant, we use a low-rank branch initialized by SVD to reduce the quantization error. This can be formulated as follows~\cite{li2025svdquant}: 
\begin{align}
XW =X L_1L_2+ XR\approx XL_1L_2+Q(X)\,Q(R),
\label{eq:svd}
\end{align}
where the SVD of $W=U\Sigma V$, the $r$ rank optimal solution is $L_1=U\Sigma_{:,:r},L_2=V_{:r,:}$, and $R=W-L_1L_2$.
However, most update rules for low-rank branches used in previous works are heuristic~\cite{li2025svdquant,guo2023lq}, and do not guarantee convergence. 
Meanwhile, existing approaches predominantly concentrate on minimizing per-layer quantization error and often neglect the final results. Moreover, previous studies~\cite{zhang2024qera,deng2023mixed} find that minimizing per-layer reconstruction error does not necessarily optimize the generation quality. We need a more suitable update scheme for the low-rank branches.

To solve this problem, a straightforward alternative is Low-Rank Adaptation (LoRA)~\cite{hu2022lora}, a Parameter-Efficient Fine-Tuning method, to align the quantized model with the FP model. However, although this approach is effective, the resulting image quality remains unsatisfactory. Inspired by Fbquant~\cite{liu2025fbquant}, we confirmed that training the SVD‑initialized low‑rank branch would regularize the model to not deviate too far from the original one. Meanwhile, this strong constraint impedes distillation from the FP model to the quantized model. To remedy this problem, we introduce additional degrees of freedom into optimization. Therefore, we can conduct effective model fine-tuning. 

Specifically, we propose Quantization-Distillation Low-Rank Adaptation (QD-LoRA). We train two low-rank branches simultaneously, while maintaining an identical number of parameters with SVDQuant~\cite{li2025svdquant}. The first branch is initialized via SVD and constrained by the backbone weights during training. It is designed to prevent excessive deviation from the original parameters. The second branch utilizes standard LoRA initialization and is dedicated to distilling knowledge from the FP model to the quantized one. The inference procedure is formulated as follows:
\begin{equation}
\begin{aligned}
XW &= X L_1L_2+XR \\
    &\approx XAB+XL_1L_2+Q(X)\,Q(R),
\end{aligned}
\label{eq:dual_lora}
\end{equation}
where we use a random Gaussian initialization for $A$ and zero for $B$ and $R=W-L_1L_2$. After training, the two low-rank branches can be merged. The auxiliary low-rank branch does not introduce any additional computational overhead during inference compared with SVDQuant~\cite{li2025svdquant}.

In practice, we find that QD-LoRA achieves superior face restoration performance compared to optimizing a single low-rank branch with an identical number of parameters.

\begin{table*}[t]
\centering
\resizebox{0.9\textwidth}{!}{
\begin{tabular}{c | c | c c c c c c c c c c }
\toprule
\rowcolor{cvprblue!30}{\shortstack[c]{Bits\\(W/A)}} & {Methods}  & CLIP-IQA$\uparrow$  & DISTS$\downarrow$ & LPIPS$\downarrow$ & MANIQA$\uparrow$ & MUSIQ$\uparrow$ & NIQE$\downarrow$ & Deg.$\downarrow$ & LMD$\downarrow$ & \shortstack[c]{FID\\(FFHQ)}$\downarrow$ & \shortstack[c]{FID\\(CelebA)}$\downarrow$\\
\midrule
32/32 & OSDFace~\cite{osdface} & 0.6910& 0.1773& 0.3365&0.5423 & 75.64& 3.884& 60.0708& 5.287& 45.41&17.06\\
\cline{1-12}

& MaxMin~\cite{jacob2017quantizationtrainingneuralnetworks} & 0.1781& 0.4022& 0.6082& 0.1406& 12.95& 11.512& 79.7129& 11.214& 234.94&216.67\\
& Q-Diffusion~\cite{qdiffusion} & 0.3679& 0.2853& 0.4784& 0.1770& 24.23& 6.821& 73.5904& 7.840& 81.43&75.69\\
& EfficientDM~\cite{He2023EfficientDM} & 0.4571& 0.2493& 0.4128& 0.3820& 53.75& 7.269&63.4590& 5.800& 63.47&34.77\\
& PassionSR~\cite{zhu2024passionsr} & 0.6309& \R{0.1762}& \B{0.3368}& 0.4731& 72.74& 4.125& 61.5067& 5.633& \B{43.08}&23.00\\
& SVDQuant~\cite{li2025svdquant} & \B{0.6566}& \B{0.1774}& 0.3397& \B{0.4818}&\B{73.59}& \B{4.021}& \B{60.7007}& \B{5.501}& 44.57&\B{22.44}\\
\rowcolor{cvprblue!10}\cellcolor{white}\multirow{-6}{*}{6/6}& QuantFace &\R{0.6670} & 0.1783& \R{0.3345}&\R{0.5122} & \R{74.89}& \R{3.759}& \R{60.5890}& \R{5.413}& \R{42.99}&\R{19.02}\\
\cline{1-12}

& MaxMin~\cite{jacob2017quantizationtrainingneuralnetworks} & 0.1694	&0.4144	&0.6354&	0.1472	&12.93&	12.864&	80.8289&	14.038&	261.82&	251.00\\
& Q-Diffusion~\cite{qdiffusion}  & 0.2399	&0.3541&	0.5505&	0.1141&	16.45&	9.091&	77.0839&9.125&	136.06&	118.48\\
& EfficientDM~\cite{He2023EfficientDM}  & 0.4981&	0.2475&	0.4113&	0.4119&	55.96&	7.212&	64.8563&	6.096&	60.11&	33.96\\
& PassionSR~\cite{zhu2024passionsr} & 0.6391	&\R{0.1803}&	\B{0.3401}&	0.4719&	\B{72.38}&	4.326&	\B{63.3448}	&\B{5.737}&	\R{45.73}&	\B{26.01}\\
& SVDQuant~\cite{li2025svdquant}  & \R{0.6912}&	0.1884&	0.3639&	\B{0.4732}&	71.79&	\R{3.762}&	63.3879&	6.043&	50.51&	33.18\\
\rowcolor{cvprblue!10}\cellcolor{white}\multirow{-6}{*}{4/6}& QuantFace   & \B{0.6660}&	\B{0.1804}&	\R{0.3329}	&\R{0.5317}	&\R{75.33}&	\B{4.128}&	\R{62.0754}&	\R{5.457}&	\B{49.59}&	\R{19.51}\\
\cline{1-12}

& MaxMin~\cite{jacob2017quantizationtrainingneuralnetworks}  &0.2045&	0.4466&	0.6867&	0.2137&	13.45&	18.011&	82.1358&	16.783&	315.62&	311.01\\
& Q-Diffusion~\cite{qdiffusion}  & 0.1747&	0.4258&	0.6660&	0.1838&	13.26&	16.620&	82.0933&	17.066&	295.58&	289.39\\
& EfficientDM~\cite{He2023EfficientDM}  &0.1992&	0.4326&	0.6368&	0.1692&	13.05&	15.275&	81.6171&	15.463&	265.51&	254.24\\
& PassionSR~\cite{zhu2024passionsr} & 0.3248&	0.3426&	0.5603&	0.1398&	15.66&	6.821&	82.4272&	18.233&	250.54&	279.35\\
& SVDQuant~\cite{li2025svdquant}  & 0.5192&	0.2563&	0.4143&	0.3572&	58.90&	5.709&	76.3504&	10.387&	84.91&	72.36\\
\rowcolor{cvprblue!10}\cellcolor{white}\multirow{-6}{*}{4/4}& QuantFace & \B{0.6916}&	\B{0.1914}	&\B{0.3709}&	\B{0.5222}&	\B{73.69}&	\R{4.072}&	\B{69.2888}&	\B{6.195}&	\B{58.34}&	\B{31.17}\\

\rowcolor{cvprblue!10}\cellcolor{white}{4/3.97}& QuantFace-MP  & \R{0.6984}&	\R{0.1869}	&\R{0.3518}&	\R{0.5560}&	\R{75.17}&	\B{4.116}&	\R{65.6640}&	\R{5.927}&	\R{53.18}&	\R{25.08}\\
\hline
\bottomrule
\end{tabular}
}
\vspace{-1mm}
\caption{Quantitative comparison on the \textit{synthetic dataset} CelebA-Test~\cite{CelebA-Test}. The best and second best results are colored with \textcolor{red}{red} and \B{blue}.}
\label{tab:synthetic}
\end{table*}

\begin{figure*}[t]
\Large
\centering
\newcommand{\imgid}{2862}
\newcommand{\imgnote}{2862}

    \begin{adjustbox}{max width=0.9\textwidth}
    \begin{tabular}{ccccccccc}
    \includegraphics[width=0.2\textwidth]{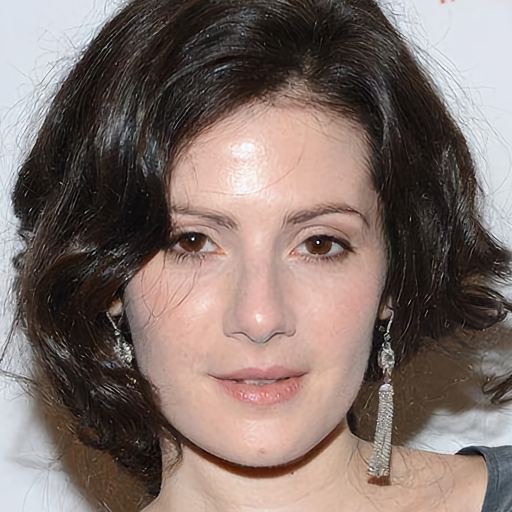}  &
    \includegraphics[width=0.2\textwidth]{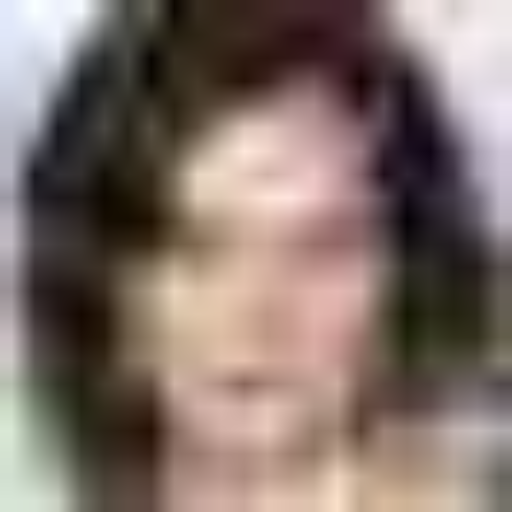}  &
    \includegraphics[width=0.2\textwidth]{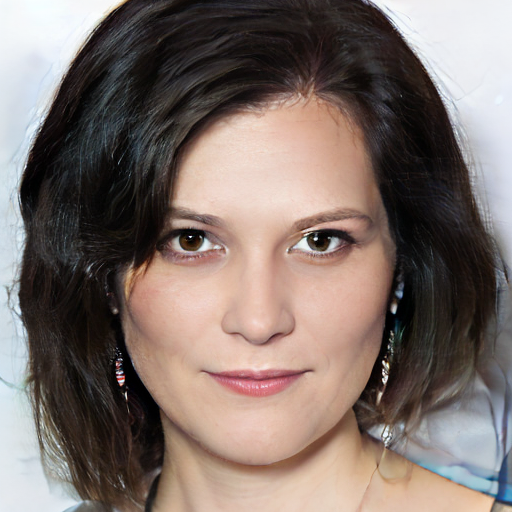}  &
    \includegraphics[width=0.2\textwidth]{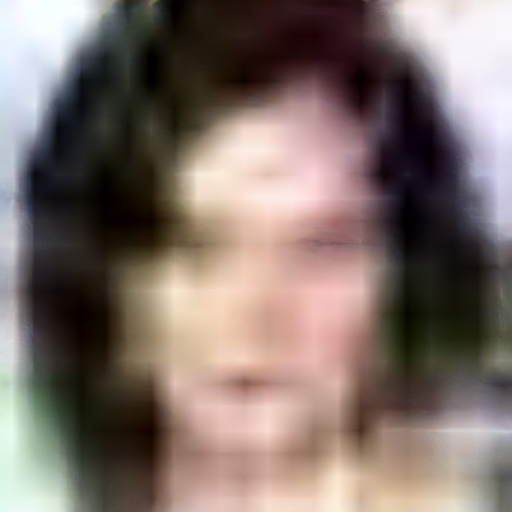} &
    \includegraphics[width=0.2\textwidth]{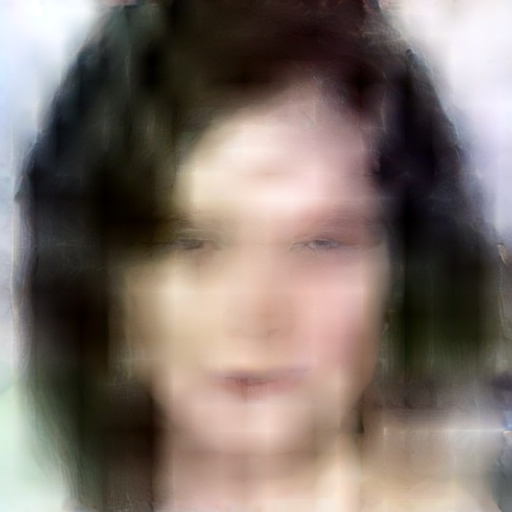}  &
    \includegraphics[width=0.2\textwidth]{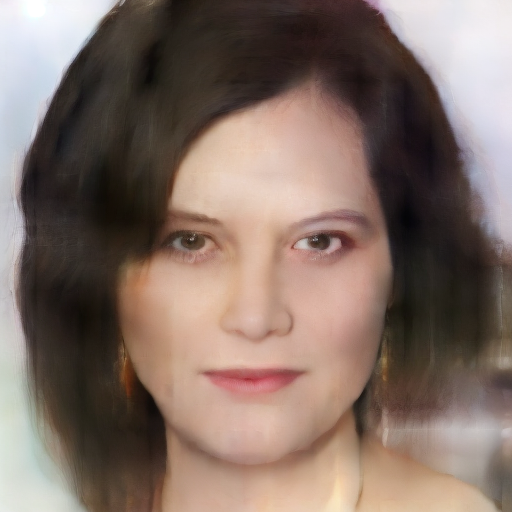}  &
    \includegraphics[width=0.2\textwidth]{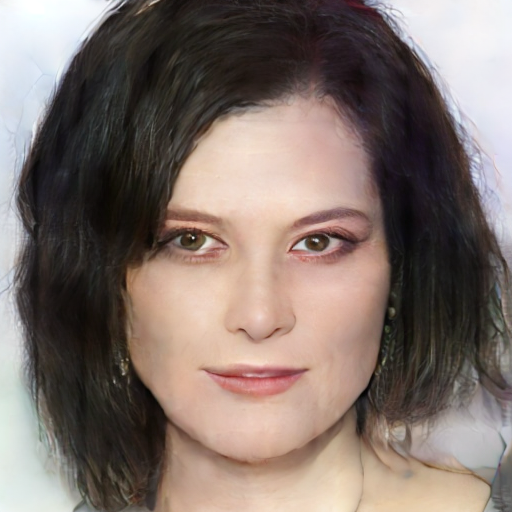}  &
    \includegraphics[width=0.2\textwidth]{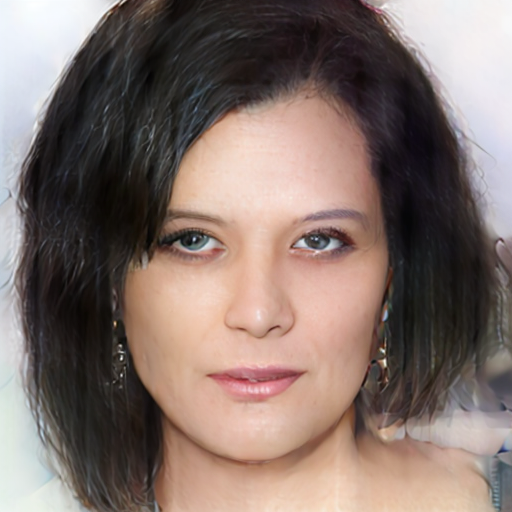} &
    \includegraphics[width=0.2\textwidth]{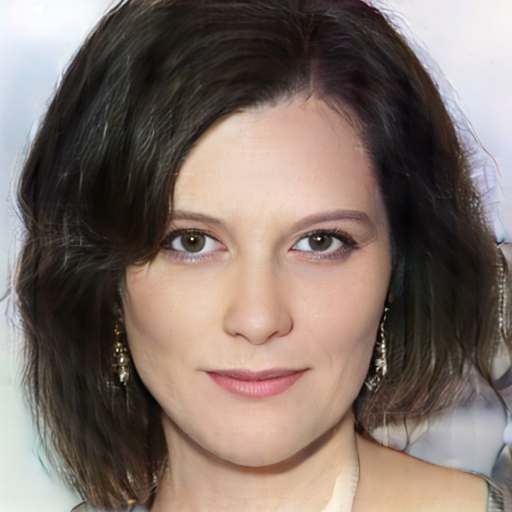} 
    \\
    HQ - \imgnote &
    LQ - \imgnote  &
    OSDFace~\cite{osdface}  &
    MinMax~\cite{jacob2017quantizationtrainingneuralnetworks} & 
    Q-Diffusion~\cite{qdiffusion}  &
    EfficientDM~\cite{He2023EfficientDM}  &
    PassionSR~\cite{zhu2024passionsr}  &
    SVDQuant~\cite{li2025svdquant} &
    QuantFace (ours) 
    \\
    \multicolumn{2}{c}{Bits (W/A)} & W32A32
    &W4A6 &W4A6 &W4A6 &W4A6 &W4A6 & W4A6 
    \\
    \end{tabular}
    \end{adjustbox}

\renewcommand{\imgid}{254}
\renewcommand{\imgnote}{0254}

    \begin{adjustbox}{max width=0.9\textwidth}
    \begin{tabular}{ccccccccc}
    \includegraphics[width=0.2\textwidth]{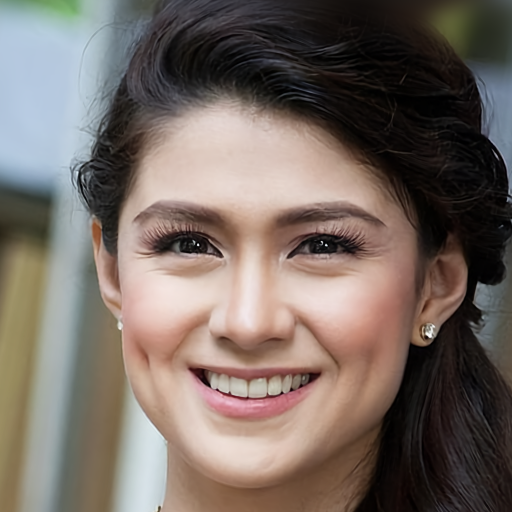}  &
    \includegraphics[width=0.2\textwidth]{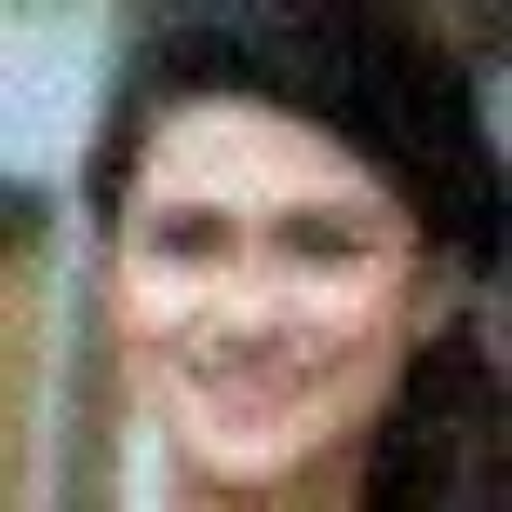}  &
    \includegraphics[width=0.2\textwidth]{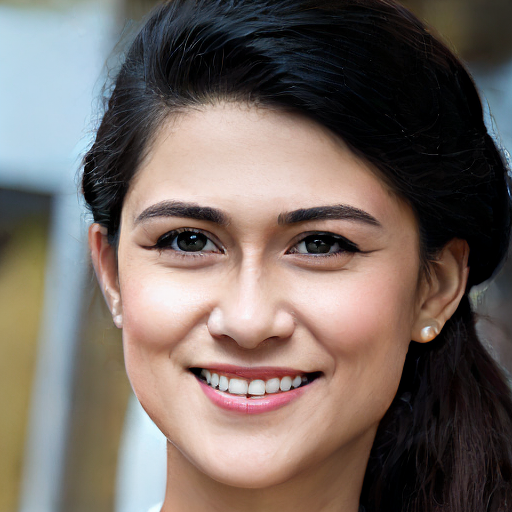}  &
    \includegraphics[width=0.2\textwidth]{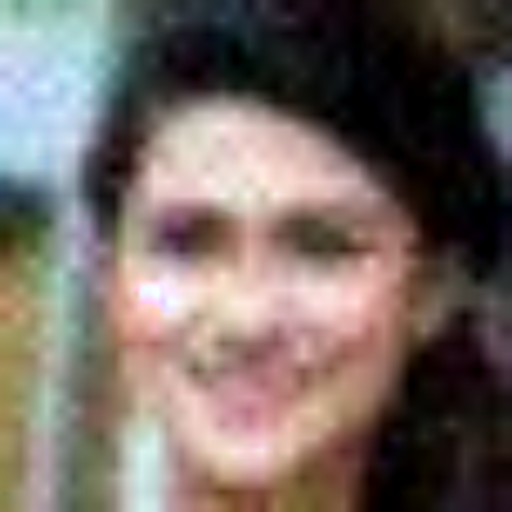} &
    \includegraphics[width=0.2\textwidth]{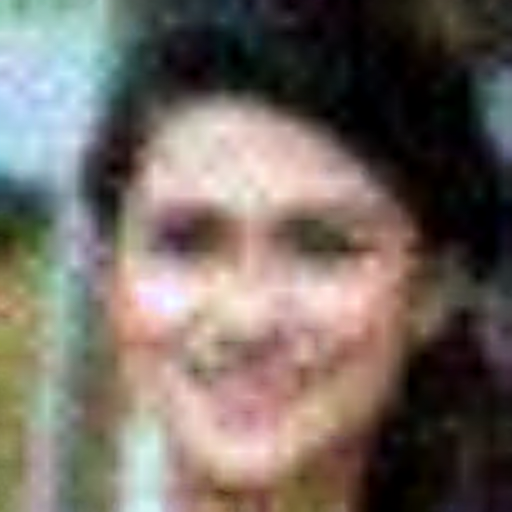}  &
    \includegraphics[width=0.2\textwidth]{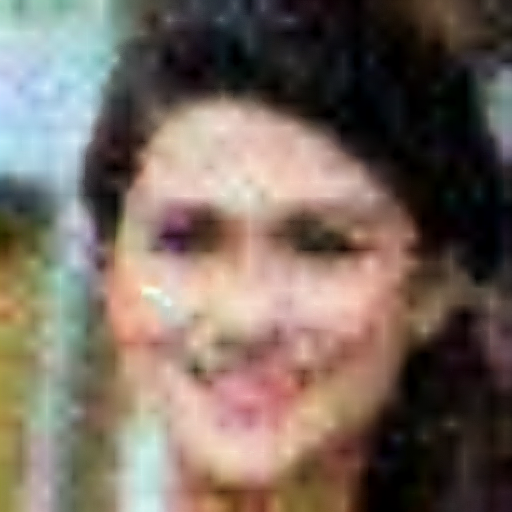}  &
    \includegraphics[width=0.2\textwidth]{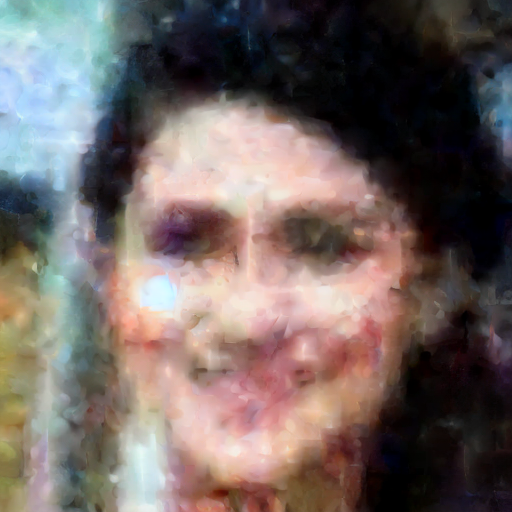}  &
    \includegraphics[width=0.2\textwidth]{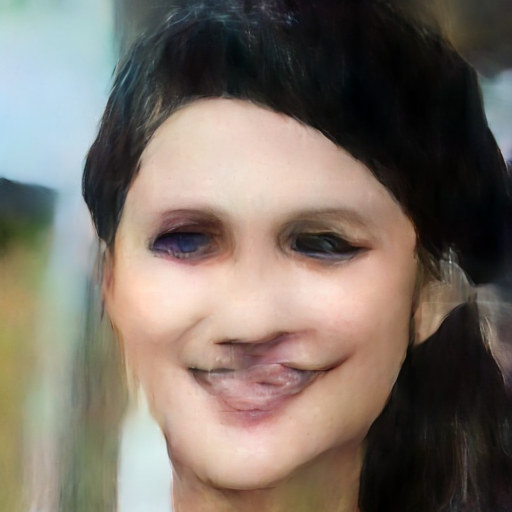} &
    \includegraphics[width=0.2\textwidth]{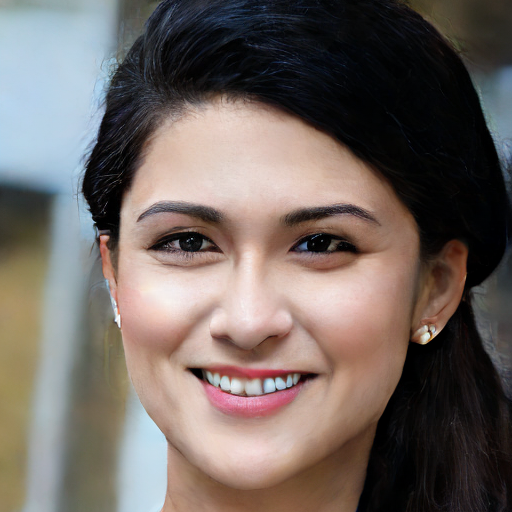} 
    \\
    HQ - \imgnote &
    LQ - \imgnote  &
    OSDFace~\cite{osdface}  &
    MinMax~\cite{jacob2017quantizationtrainingneuralnetworks} & 
    Q-Diffusion~\cite{qdiffusion}  &
    EfficientDM~\cite{He2023EfficientDM}  &
    PassionSR~\cite{zhu2024passionsr}  &
    SVDQuant~\cite{li2025svdquant} &
    QuantFace (ours) 
    \\
    \multicolumn{2}{c}{Bits (W/A)} & W32A32
    &W4A4 &W4A4 &W4A4 &W4A4 &W4A4& W4A3.97
    \\
    \end{tabular}
    \end{adjustbox}

\renewcommand{\imgid}{0969}
\renewcommand{\imgnote}{0969}
    \begin{adjustbox}{max width=0.9\textwidth}
    \begin{tabular}{ccccccccc}
    \includegraphics[width=0.2\textwidth]{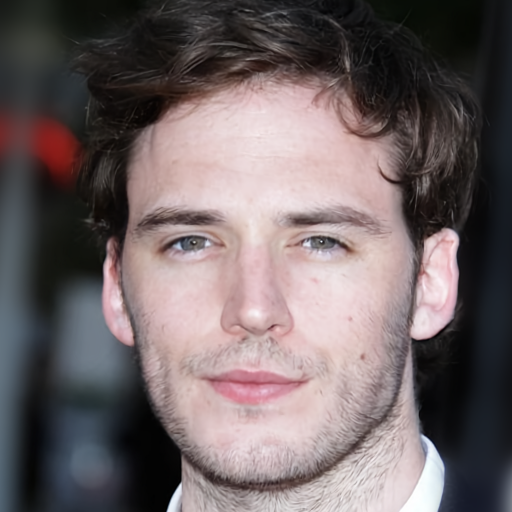}  &
    \includegraphics[width=0.2\textwidth]{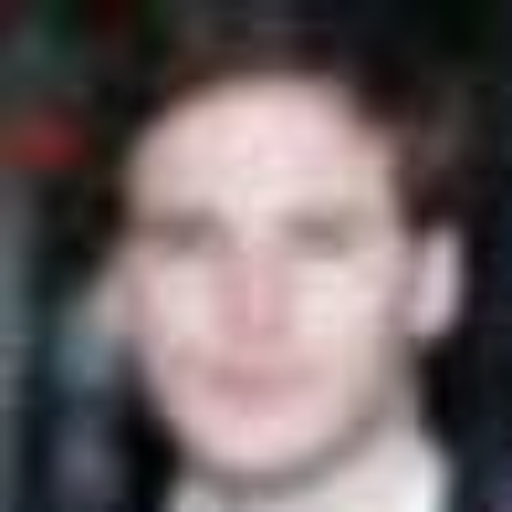}  &
    \includegraphics[width=0.2\textwidth]{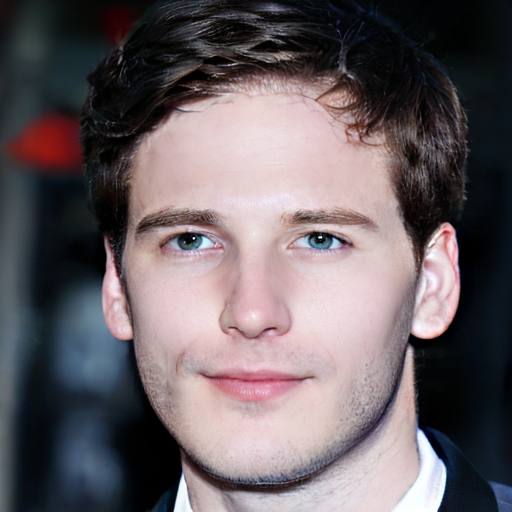}  &
    \includegraphics[width=0.2\textwidth]{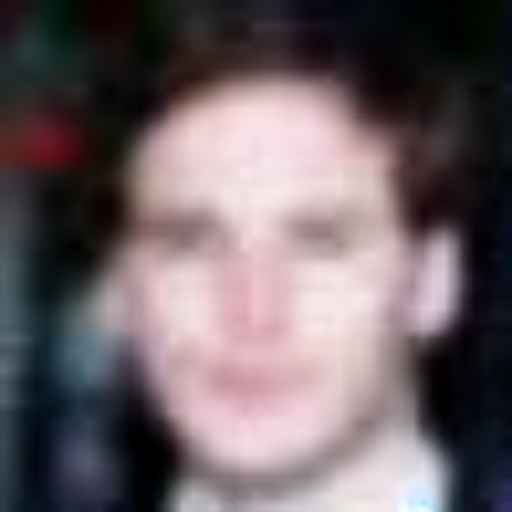} &
    \includegraphics[width=0.2\textwidth]{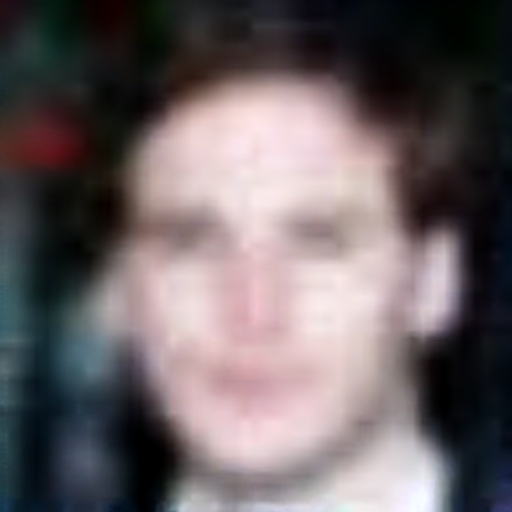}  &
    \includegraphics[width=0.2\textwidth]{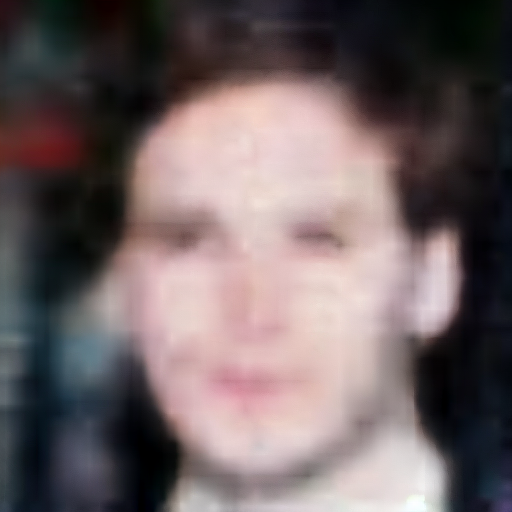}  &
    \includegraphics[width=0.2\textwidth]{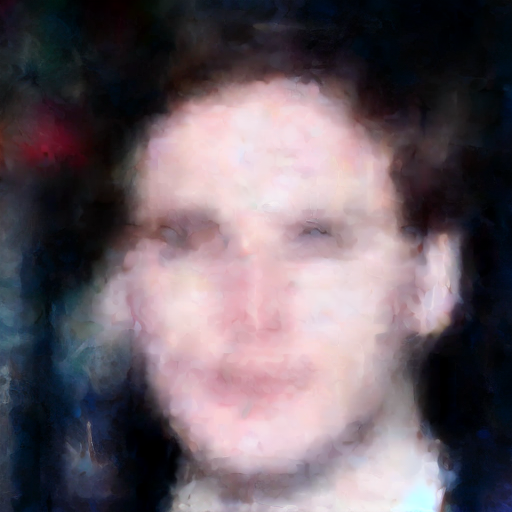}  &
    \includegraphics[width=0.2\textwidth]{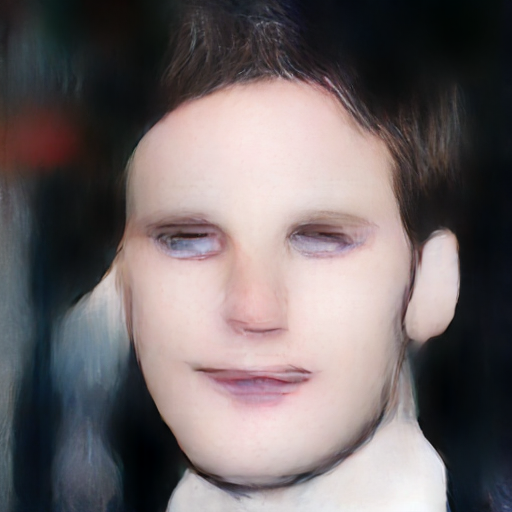} &
    \includegraphics[width=0.2\textwidth]{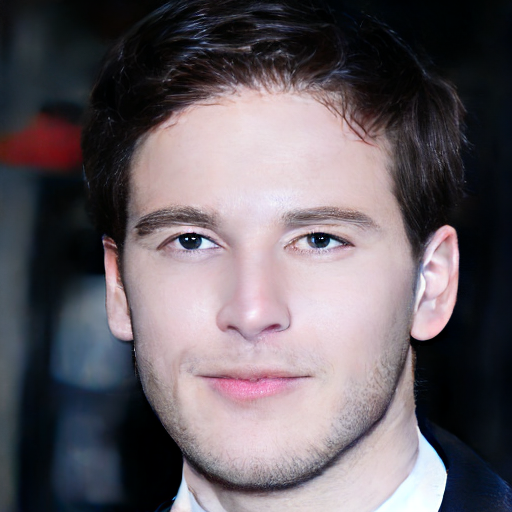} 
    \\
    HQ - \imgnote &
    LQ - \imgnote  &
    OSDFace~\cite{osdface}  &
    MinMax~\cite{jacob2017quantizationtrainingneuralnetworks} & 
    Q-Diffusion~\cite{qdiffusion}  &
    EfficientDM~\cite{He2023EfficientDM}  &
    PassionSR~\cite{zhu2024passionsr}  &
    SVDQuant~\cite{li2025svdquant} &
    QuantFace (ours) 
    \\
    \multicolumn{2}{c}{Bits (W/A)} & W32A32
    &W4A4 &W4A4 &W4A4 &W4A4 &W4A4 & W4A3.97 
    \\
    \end{tabular}
    \end{adjustbox}
\vspace{-3mm}
\caption{Visual comparison of the \textit{synthetic} CelebA-Test dataset in challenging cases.}
\label{fig:vis-celeba}
\vspace{-4mm}
\end{figure*}
\subsection{Adaptive Bit-width Allocation}
\vspace{-2mm}
Another challenge in low-bit quantization is the suboptimal allocation of limited resources. Previous works~\cite{sui2024bitsfusion} notice that different layers exhibit varying sensitivities to quantization. We measure the change in FID~\cite{fid} on FFHQ~\cite{ffhq} by quantizing the activations of each layer individually, and we observe analogous findings in activation quantization.

As illustrated in Fig.~\ref{fig:fid_sensitivity}, within the OSDFace UNet, key convolution layers and residual connections have an outsized impact on the final image quality. After identifying the quantization bottlenecks, we can then devise a mixed‑precision scheme. Enlightened by prior works~\cite{zhao2024mixdq, sui2024bitsfusion, ma2023ompq}, we model the bit-width allocation as an integer programming problem, which is called adaptive bit-width allocation. 

When designing our integer programming objective, we focus on both reconstruction error and perception metric. Prior works~\cite{vidit,sui2024bitsfusion} observe that simply allocating higher bit‑widths to layers with large reconstruction error (\ie, MSE) does not guarantee better visual performance. Consequently, we employ a heuristic approach to assign perceptual error weight to each activation layer. Specifically, we first quantize the activation of every single layer to 4-bit. We then progressively increase the bit-width of the quantized activation until the FID difference compared to the FP model is within $\epsilon=0.02$, or until the maximum bit-width $B_{max}=8$ is reached. Then, for each layer $i$, we extract the corresponding bit-width $B_i$ and assign the weight as $w_i=2^{B_i}-1$.

Given the total bits budget $\mathcal{B}$, and candidate bit-widths $b\in\{3,4,5,6\}$, the quantization error of each layer is $\mathcal{L}_{i,b}=w_i^2\left \|WX-Q(W)Q(X) \right \|_F^2$. After collecting the weighted quantization errors for each layer, the optimization problem can be solved in tens of seconds to a few minutes. 

The Detailed integer programming workflow can be formulated as follows~\cite{zhao2024mixdq}:
\begin{equation}
    \begin{aligned}
    &\underset{a_{i,b}}{\text{argmin}} \quad \sum_{i=1}^{N} \sum_{b=3,4,5,6} a_{i,b} \cdot \mathcal{L}_{i, b}  \\
    &\text{s.t.} \quad \sum_{b=3,4,5,6} a_{i,b} = 1, \quad \sum_{i=1}^{N} \sum_{b=3,4,5,6} a_{i,b} \cdot \mathcal{M}_{i,b} \leq \mathcal{B}, \\
    &\quad a_{i,b} \in \{0,1\}, \quad \forall i \in \{1,\cdots,N\}, \forall b \in \{3,4,5,6\},
    \end{aligned}
\label{equ:integer-planning}
\end{equation}
where $N$ is the number of layers, $a_{i,b}=1$ denotes that the $i$-th layer will be quantized to $b$-bit, and $\mathcal{M}_{i,b}$ indicates the cost of the $i$-th layer when it is quantized to $b$-bit.

Under the overall W4A4 resource constraints, we applied the aforementioned integer programming approach to allocate appropriate bit-widths for activation quantization.
\section{Experiments}
\label{sec:experiments}
\subsection{Experiment Settings}
\vspace{-2mm}
\paragraph{Data Construction.}We randomly select 1000 high-quality face images from FFHQ~\cite{ffhq} and resized them to 512$\times$512 pixels. We obtained our synthetic training data using a dual-stage degradation model, which is consistent with WaveFace~\cite{waveface}. 
We evaluate our quantized model on both synthetic and real‑world datasets.  The synthetic dataset is CelebA-Test~\cite{CelebA-Test} from DAEFR~\cite{daefr}, and the real-world datasets are Wider-Test~\cite{wider-test}, WebPhoto-Test~\cite{webphoto-test}, and LFW-Test~\cite{lfw-test}, which are consistent with OSDFace~\cite{osdface}.
\vspace{-5mm}
\paragraph{Metrics.} For the \textit{Synthetic Datasets}, LPIPS~\cite{lpips} and DISTS~\cite{dists} are adopted as reference-based perceptual measures, together with CLIPIQA~\cite{clipiqa}, MANIQA~\cite{maniqa}, NIQE~\cite{niqe}, and MUSIQ~\cite{musiq} as non-reference metrics. FID~\cite{fid} with both FFHQ~\cite{ffhq} and CelebA-Test HQ~\cite{CelebA-Test} are used to evaluate the distribution similarity between the real faces and generated ones. Moreover, following the previous face restoration works~\cite{vqfr,wider-test,daefr},  we also evaluate the embedding angle of ArcFace~\cite{arcface} called `Deg.', and the landmark distance named `LMD'. For the \textit{Real-world Datasets}, CLIPIQA~\cite{clipiqa}, MANIQA~\cite{maniqa}, MUSIQ~\cite{musiq}, NIQE~\cite{niqe}, and FID~\cite{fid} with FFHQ~\cite{ffhq} are employed.
\vspace{-5.5mm}
\paragraph{Implementation Details.} Learning rate for QuantFace is set to $10^{-5}$ using the Adam~\cite{adam} optimizer. Only the low-rank branches of each layer are updated during training, and the LoRA rank for both branches is 16. The overall loss function for the generator is defined as
\begin{equation}
\begin{aligned}
\mathcal{L}_\text{gen} &= \lambda_\text{dis} \cdot \mathcal{L}_\mathcal{G}(z_q, z_{fp}) 
                      + \lambda_\text{ID} \cdot \mathcal{L}_\text{ID}(I_q, I_{fp}) \\
                      &+ \lambda_\text{per} \cdot \mathcal{L}_{\text{LPIPS}}(I_q, I_{fp}) 
                      + \lambda_{mse} \cdot \operatorname{MSE}(I_q, I_{fp}).
\end{aligned}
\end{equation}
To keep the face identity, we utilize a pretrained ArcFace model~\cite{arcface} to encode both the quantized and FP faces into identity embeddings, where
\begin{equation}
\mathcal{L}_\text{ID}  = 1 - \cos\left( \mathcal{F}(I_q), \mathcal{F}(I_{fp}) \right).
\end{equation}
Furthermore, to enhance the quality of quantized faces, we adopt the same GAN training strategy as that employed in OSDFace~\cite{osdface}, where
\begin{equation}
\begin{aligned}
\mathcal{L}_\mathcal{G} &= -\mathbb{E}_t \left[\log \mathcal{D}_{\theta} \left( F(z_q, t) \right)\right],\\
\mathcal{L}_\mathcal{D} &= -\mathbb{E}_t \left[\log \left( 1 - \mathcal{D}_{\theta} \left( F(z_q, t) \right) \right)\right] \\
&- \mathbb{E}_t \left[\log \mathcal{D}_{\theta} \left( F(z_{fp}, t) \right)\right], 
\end{aligned}
\end{equation}
Experiments are conducted on a single NVIDIA RTX A6000 GPU, consuming 29 GB of GPU memory and 6 hours of GPU time for training.
\vspace{-5.5mm}
\paragraph{Compare Methods.} We compare QuantFace with several representative quantization methods: MaxMin~\cite{jacob2017quantizationtrainingneuralnetworks}, Q-Diffusion~\cite{qdiffusion}, EfficientDM~\cite{He2023EfficientDM}, PassionSR~\cite{zhu2024passionsr}, and SVDQuant~\cite{li2025svdquant}. Q-Diffusion~\cite{qdiffusion}, EfficientDM~\cite{He2023EfficientDM}, and SVDQuant~\cite{li2025svdquant} are quantization methods for multi-step diffusion models. PassionSR~\cite{zhu2024passionsr} is a quantization method designed for one-step diffusion models.
\begin{table}
\centering
\resizebox{0.8\linewidth}{!}{
\setlength{\tabcolsep}{1mm}
    \begin{tabular}{c | c | c c}
        \toprule
        \rowcolor{cvprblue!30}
        {Method} & {Bits} & {Params / M} ($\downarrow$ Ratio)  & {Ops / G} ($\downarrow$ Ratio)\\
        \midrule
        OSDFace~\cite{osdface}  & W32A32 & 866 ($\downarrow$0\%) & 678 ($\downarrow$0\%) \\
        \midrule
        \rowcolor{cvprblue!10}
        & W6A6 & 185 ($\downarrow$78.61\%) & 158 ($\downarrow$76.66\%) \\
        \rowcolor{cvprblue!10}
        & W4A6 & 131 ($\downarrow$84.85\%) & 135 ($\downarrow$80.10\%) \\
        \rowcolor{cvprblue!10}
        \multirow{-3}{*}{QuantFace}
        & W4A4 & 131 ($\downarrow$84.85\%) & 116 ($\downarrow$82.91\%) \\
        \bottomrule
    \end{tabular}%
}
\caption{Params, Ops, and compression ratio (UNet only) of various settings. Ops are computed with input size 512$\times$512.}
\label{tab:complexity}
\vspace{-2.5mm}
\end{table}
\vspace{-6mm}
\paragraph{Compression Ratio.} To evaluate computational cost, following previous work~\cite{qin2023quantsr}, we measure the total model size (Params / M) and the number of operations (Ops / G). The results are summarized in Tab.~\ref{tab:complexity}. Under the 4-bit setting, our QuantFace achieves approximately 84.85\% compression ratio for parameters and 82.91\%  compression ratio for computation, compared with the original UNet.

\begin{table*}[t]
\centering
\vspace{-1mm}
\setlength{\tabcolsep}{0.4mm}

\newcolumntype{?}{!{\vrule width 1pt}}
\newcolumntype{C}{>{\centering\arraybackslash}X}
\begin{adjustbox}{max width=0.9\textwidth}

\begin{tabular}{c|c|*{5}{c}|*{5}{c}|*{5}{c}}
\toprule
\rowcolor{cvprblue!30} && \multicolumn{5}{c|}{WebPhoto-Test~\cite{webphoto-test}} & \multicolumn{5}{c|}{LFW-Test~\cite{lfw-test}} & \multicolumn{5}{c}{Wider-Test~\cite{wider-test}} \\
\rowcolor{cvprblue!30}
\multirow{-2}{*}{\shortstack[c]{Bits\\(W/A)}}&\multirow{-2}{*}{Methods} & C-IQA$\uparrow$ & M-IQA$\uparrow$ & MUSIQ$\uparrow$ & NIQE$\downarrow$ & FID$\downarrow$ & C-IQA$\uparrow$ & M-IQA$\uparrow$ & MUSIQ$\uparrow$ & NIQE$\downarrow$ & FID$\downarrow$ & C-IQA$\uparrow$ & M-IQA$\uparrow$ & MUSIQ$\uparrow$ & NIQE$\downarrow$ & FID$\downarrow$ \\
\midrule
32/32&OSDFace~\cite{osdface}& 
 0.7106& 0.5162& 73.94&  3.986 &84.60
& 0.7203&0.5493 & 75.35& 3.871&44.63
&0.7284&0.5229 &74.60 & 3.774&34.65\\
\hline

\multirow{6}{*}{6/6}&MaxMin~\cite{jacob2017quantizationtrainingneuralnetworks} & 0.2263
&0.1291 &24.68 &7.816 &137.31 &0.2652
&0.1927 &39.63 &6.753 &118.28 &0.2053
&0.1062 &15.67 & 9.418&152.62   \\

&Q-Diffusion~\cite{qdiffusion} & 0.3775
&0.2145 &39.18 &5.683 &100.06 &0.4743
&0.2853 &52.38 &5.134 &72.50 &0.3741
&0.1751 &29.01 & 6.322&66.50   \\

&EfficientDM~\cite{He2023EfficientDM} & 0.5106
&0.3570 &56.85 &5.665 &100.04 &0.5793
&0.3980 &62.62 &4.818 &68.69 &0.5127
&0.3643 &52.52 &6.325 &49.09  \\

&PassionSR~\cite{zhu2024passionsr} &0.6666 
&0.4535 &70.37 &4.304 &\B{82.02} &0.6824
&0.4849 &72.94 &3.988 & 47.90&0.6733
&\B{0.4594} &70.93 &\B{4.145} &\B{39.32}   \\

&SVDQuant~\cite{li2025svdquant} & \B{0.6927}
&\B{0.4668} & \B{71.51}&\B{4.178} &84.29 &\B{0.7024}
&\B{0.4963}&\B{73.96} &\B{3.964} &\B{46.22} &\B{0.6841}
&0.4520 &\B{71.79} &4.171 &39.38   \\

\rowcolor{cvprblue!10}\cellcolor{white}&QuantFace &\R{0.7040} 
&\R{0.4953} &\R{73.26} &\R{3.837} &\R{80.74} 
&\R{0.7119} &\R{0.5292} &\R{74.88} &\R{3.810} &\R{46.19} 
&\R{0.7115}&\R{0.4984} &\R{73.88} &\R{3.685} & \R{37.57}  \\
\hline

\multirow{6}{*}{4/6}&MaxMin~\cite{jacob2017quantizationtrainingneuralnetworks} & 0.2204
& 0.1246&22.47 & 8.565& 143.13&0.2538
&0.1924 &36.46 &7.266 &121.44 &0.1978
&0.1027 &14.75 &10.228 &165.05   \\

&Q-Diffusion~\cite{qdiffusion} & 0.2908
&0.1568 &28.86 &6.841 &116.86 &0.3558
&0.2193 &42.46 &6.027 & 94.05&0.2552
&0.1132 & 19.27&7.918 &104.15   \\

&EfficientDM~\cite{He2023EfficientDM} & 0.5840
&0.4120 &61.48 &5.442 &92.52 &0.6166
&0.4314 &64.51 &4.716 &63.82 &0.5479
&0.3960 &54.76 &6.292 &47.06  \\

&PassionSR~\cite{zhu2024passionsr} & \B{0.6578}
&\B{0.4423} &69.01 &\B{4.456} &83.98 &\B{0.6485}
&\B{0.4501} &70.58 &\B{4.205} &52.42 &\B{0.6623}
&\B{0.4480} &69.31 &\B{4.407} & \B{41.10}  \\

&SVDQuant~\cite{li2025svdquant} & 0.6137
&0.4294 &\B{70.62} &4.597 &\R{83.95} &0.6292
&0.4368 &\B{71.78} &4.334 &\B{49.95} &0.6123
&0.4288 &\B{70.38} &4.608 &41.45  \\

\rowcolor{cvprblue!10}\cellcolor{white}&QuantFace & \R{0.6944} & \R{0.5117}&\R{73.55} &\R{4.368} &\B{87.27} 
&\R{0.6993} &\R{0.5263} &\R{74.59} &\R{4.122} &\R{46.90} 
&\R{0.7046} &\R{0.5125} &\R{73.93} &\R{4.115} &\R{37.74}   \\
\hline

\multirow{7}{*}{4/4}&MaxMin~\cite{jacob2017quantizationtrainingneuralnetworks} & 0.2281
&0.1600 &18.89 &11.055 &167.46 &0.2533
&0.2019 &28.03 &9.356 &141.90 &0.2115
&0.1497 &14.20 &12.837 &194.46 \\

&Q-Diffusion~\cite{qdiffusion} & 0.2158
&0.1328 &19.55 &10.224 &155.27 &0.2417
&0.1827 &30.42 &8.532 &128.19 &0.2010
&0.1262 &14.09 &12.239 &183.76 \\

&EfficientDM~\cite{He2023EfficientDM} & 0.2207
&0.1128 &21.43 &8.996 &161.68 &0.2589
&0.1228 &32.36 &6.779 &129.42 &0.2086
&0.1234 &14.81 &11.494 &172.48 \\

&PassionSR~\cite{zhu2024passionsr} & 0.3362
&0.1667 &27.39 &6.059 &204.05 &0.3657
&0.1857 &36.75 &6.649 &151.02 &0.3845
&0.1747 & 20.18& 6.419& 270.94  \\

&SVDQuant~\cite{li2025svdquant} & 0.5647
&0.3940 &62.98 &5.320 &111.49 &0.5741
&0.3955 &62.97 &5.023 &67.15 &0.5383
&0.3657 &58.45 &5.683 &72.41 \\

\rowcolor{cvprblue!10}\cellcolor{white}&QuantFace & \B{0.7054}&\B{0.4853} &\B{72.32} &\R{4.106} &\B{92.30} &\B{0.6861}&\B{0.4908} &\B{72.70} &\R{3.831} &\B{58.77} &\B{0.7044}&\B{0.4887} & \B{71.70}&\R{4.088}&\B{47.10} \\

\rowcolor{cvprblue!10}\cellcolor{white}4/3.97&QuantFace-MP & \R{0.7180}&\R{0.5297} &\R{73.63} &\B{4.386} &\R{91.78} &\R{0.7118}&\R{0.5426} &\R{74.66} &\B{4.111} &\R{49.57} &\R{0.7189}&\R{0.5263} & \R{73.69}& \B{4.156}&\R{40.74} \\

\hline
\bottomrule
\end{tabular}
\end{adjustbox}
\vspace{-3.5mm}
\caption{Quantitative comparison on \textit{real-world datasets}. C-IQA stands for CLIPIQA, and M-IQA stands for MANIQA. The best and second best results are colored with \textcolor{red}{red} and \B{blue}. MP stands for mixed-precision configuration generated by adaptive bit-width allocation.}
\label{table:realworld}
\vspace{-3mm}
\end{table*}
\begin{figure*}[t]
\Large
\centering
\newcommand{\imgid}{0022}
\newcommand{\imgnote}{0022}
 
    \begin{adjustbox}{max width=0.9\textwidth}
    \begin{tabular}{cccccccc}
    \includegraphics[width=0.2\textwidth]{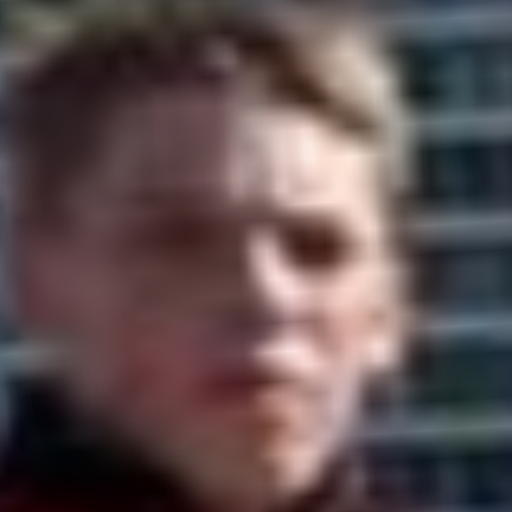}  &
    \includegraphics[width=0.2\textwidth]{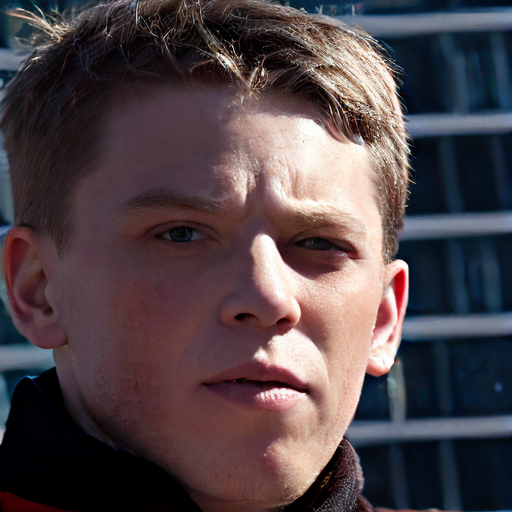}  &
    \includegraphics[width=0.2\textwidth]{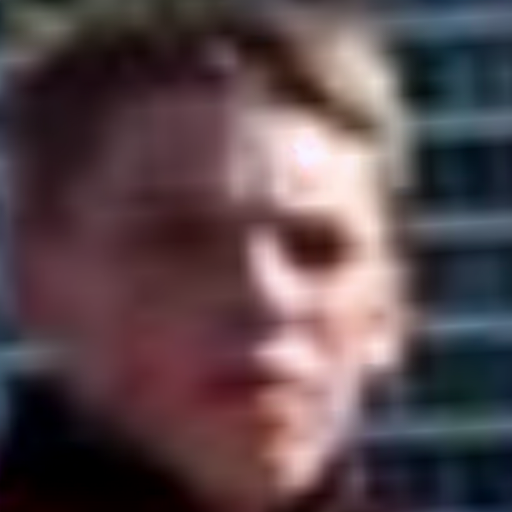} &
    \includegraphics[width=0.2\textwidth]{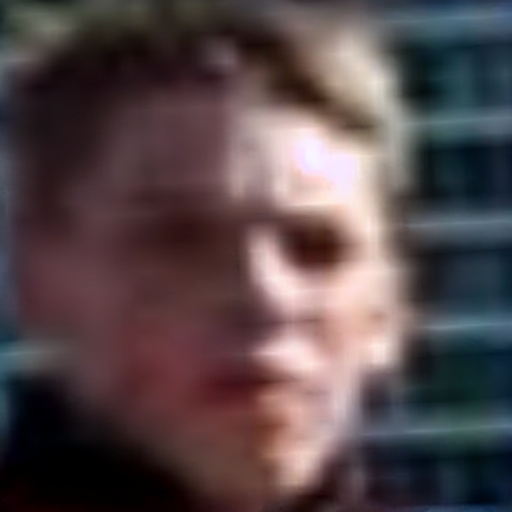}  &
    \includegraphics[width=0.2\textwidth]{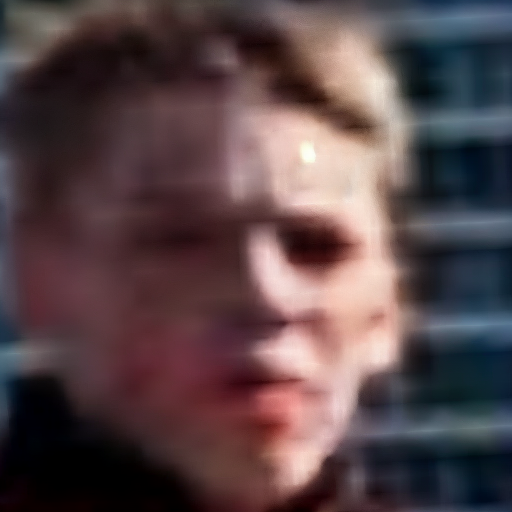}  &
    \includegraphics[width=0.2\textwidth]{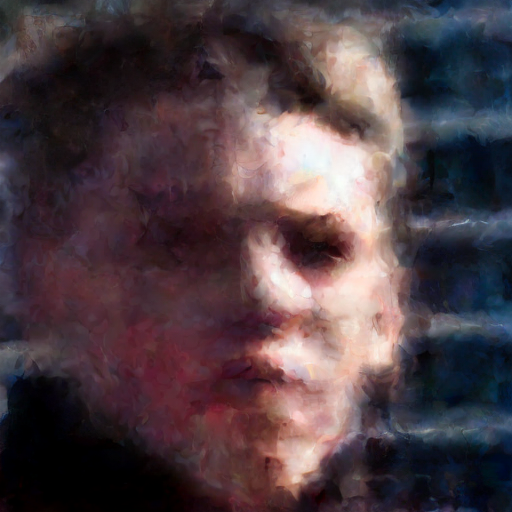}  &
    \includegraphics[width=0.2\textwidth]{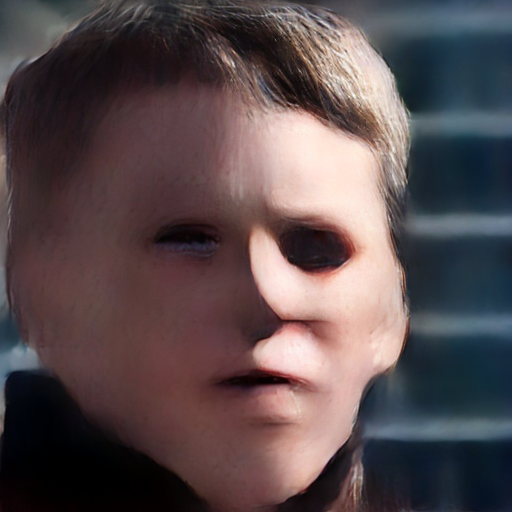} &
    \includegraphics[width=0.2\textwidth]{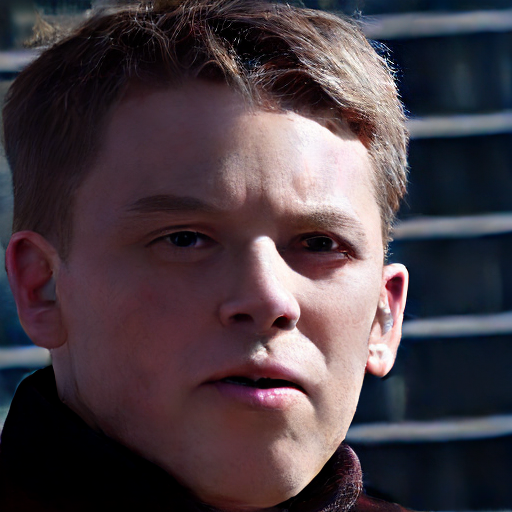} 
    \\
    Wider \imgnote  &
    OSDFace~\cite{osdface}  &
    MinMax~\cite{jacob2017quantizationtrainingneuralnetworks}  & 
    Q-Diffusion~\cite{qdiffusion}  &
    EfficientDM~\cite{He2023EfficientDM}  &
    PassionSR~\cite{zhu2024passionsr}  &
    SVDQuant~\cite{li2025svdquant} &
    QuantFace (ours) 
    \\
    \end{tabular}
    \end{adjustbox}

\renewcommand{\imgid}{00038_00}
\renewcommand{\imgnote}{00038}

    \begin{adjustbox}{max width=0.9\textwidth}
    \begin{tabular}{cccccccc}
    \includegraphics[width=0.2\textwidth]{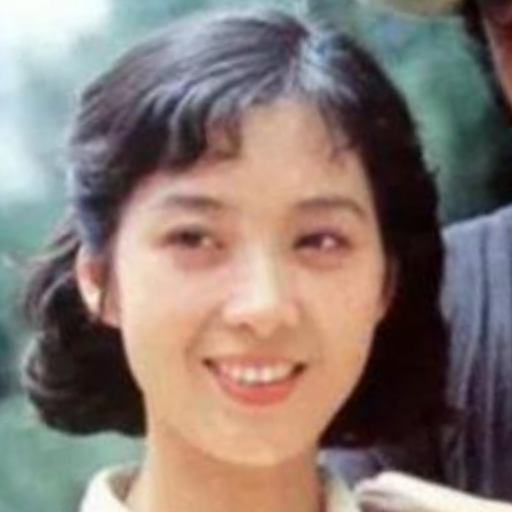}  &
    \includegraphics[width=0.2\textwidth]{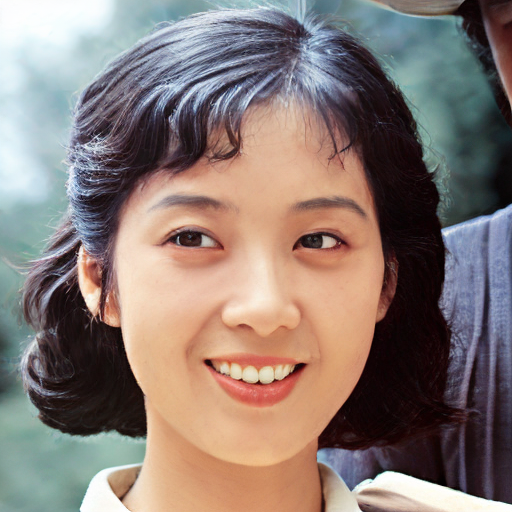}  &
    \includegraphics[width=0.2\textwidth]{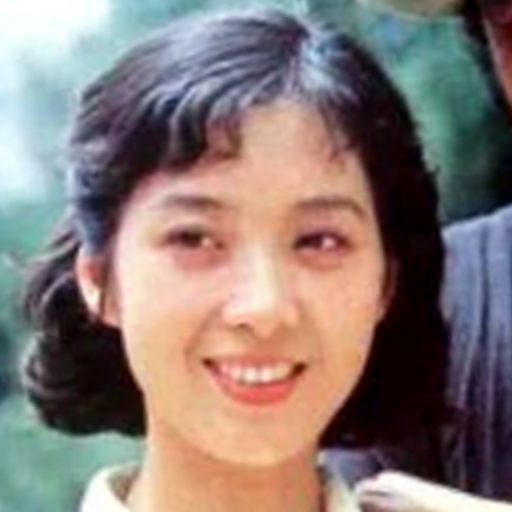} &
    \includegraphics[width=0.2\textwidth]{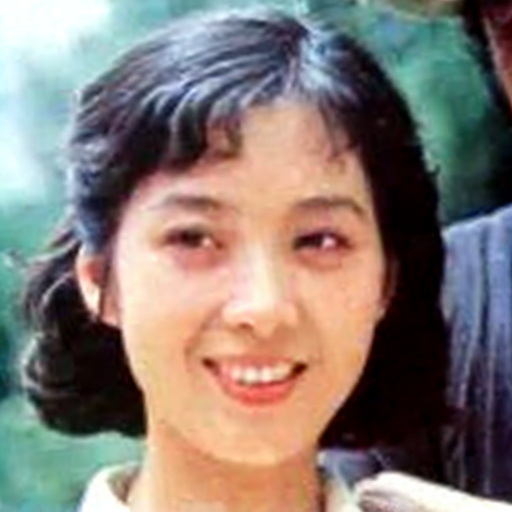}  &
    \includegraphics[width=0.2\textwidth]{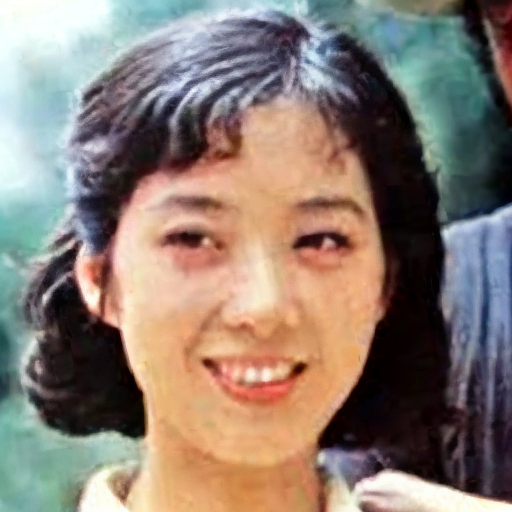}  &
    \includegraphics[width=0.2\textwidth]{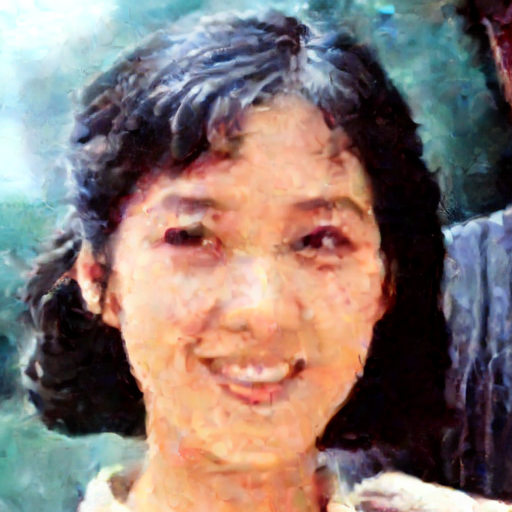}  &
    \includegraphics[width=0.2\textwidth]{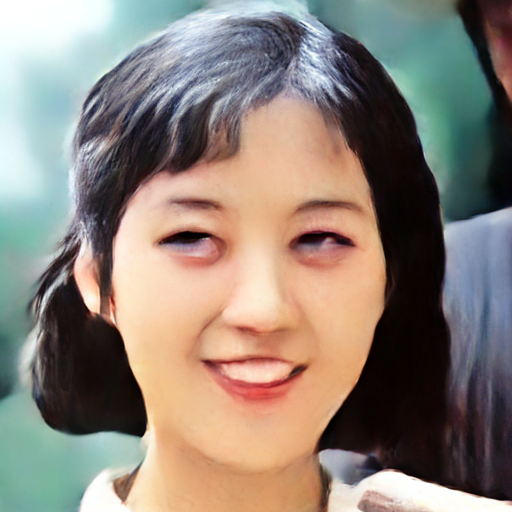} &
    \includegraphics[width=0.2\textwidth]{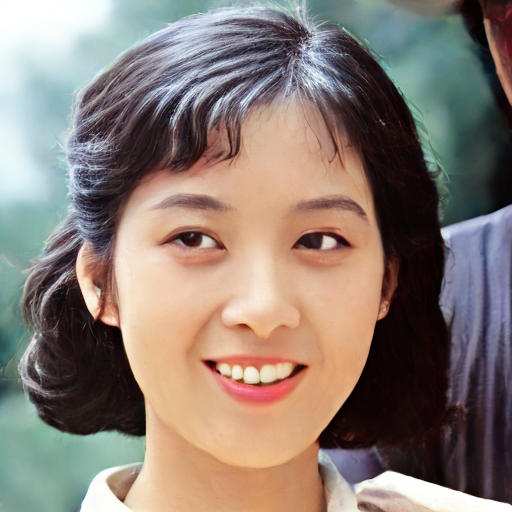} 
    \\
    WebPhoto \imgnote  &
    OSDFace~\cite{osdface}  &
    MinMax~\cite{jacob2017quantizationtrainingneuralnetworks}  & 
    Q-Diffusion~\cite{qdiffusion}  &
    EfficientDM~\cite{He2023EfficientDM}  &
    PassionSR~\cite{zhu2024passionsr}  &
    SVDQuant~\cite{li2025svdquant} &
    QuantFace (ours) 
    \\
    \end{tabular}
    \end{adjustbox}

\renewcommand{\imgid}{Elin_Nordegren}
\renewcommand{\imgnote}{E. Nordegren}

    \begin{adjustbox}{max width=0.9\textwidth}
    \begin{tabular}{cccccccc}
    \includegraphics[width=0.2\textwidth]{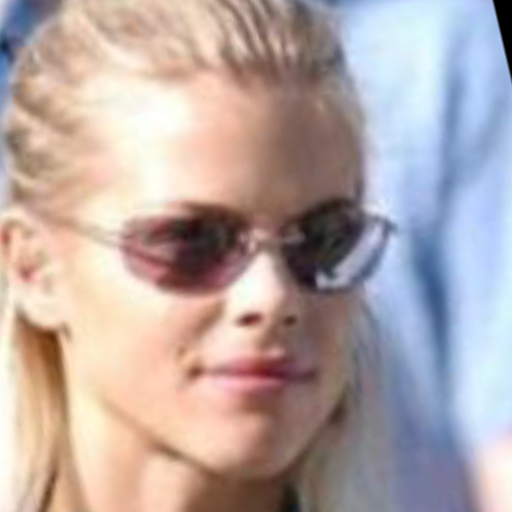}  &
    \includegraphics[width=0.2\textwidth]{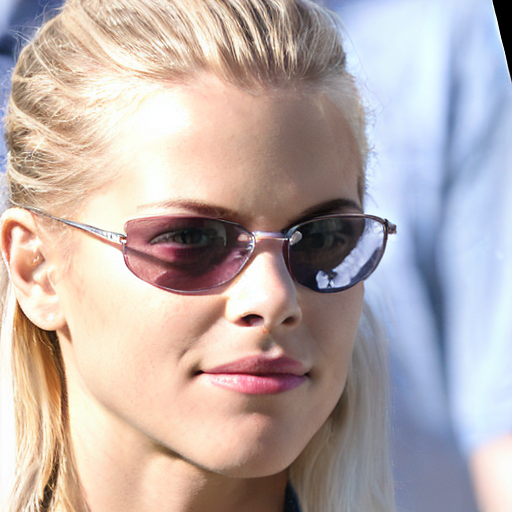}  &
    \includegraphics[width=0.2\textwidth]{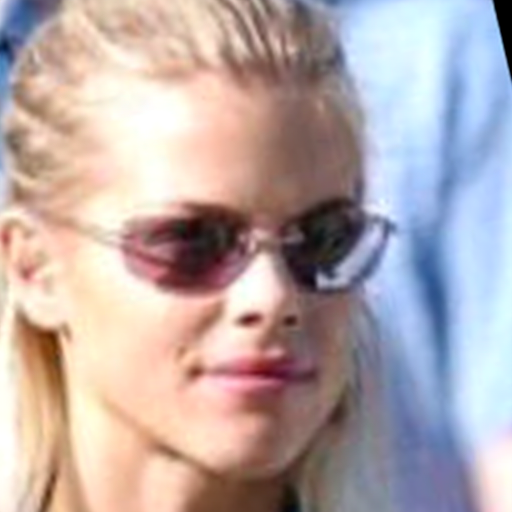} &
    \includegraphics[width=0.2\textwidth]{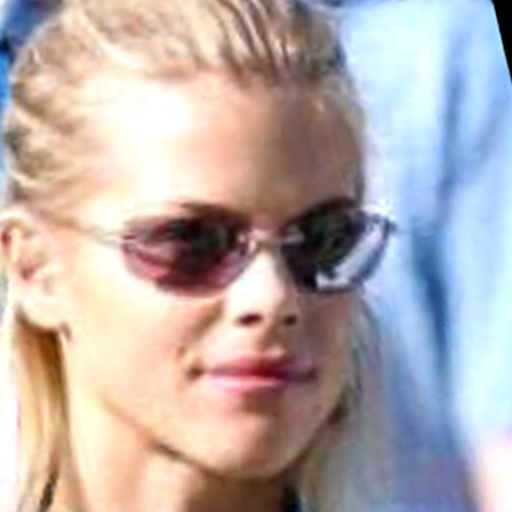}  &
    \includegraphics[width=0.2\textwidth]{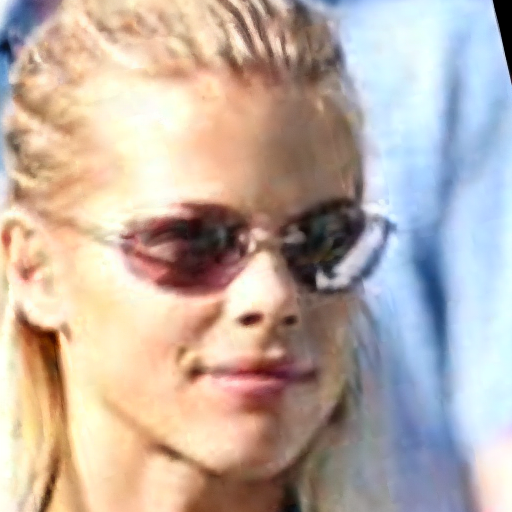}  &
    \includegraphics[width=0.2\textwidth]{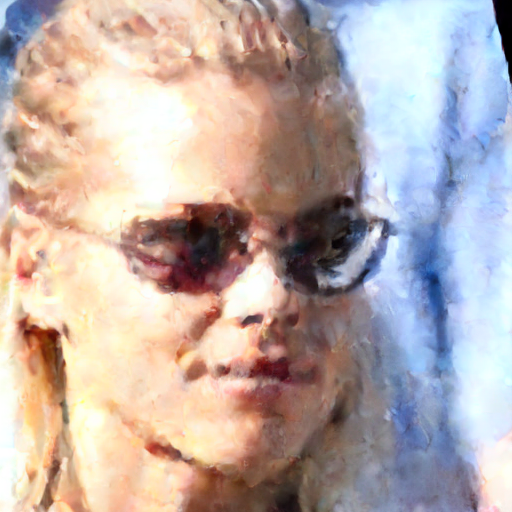}  &
    \includegraphics[width=0.2\textwidth]{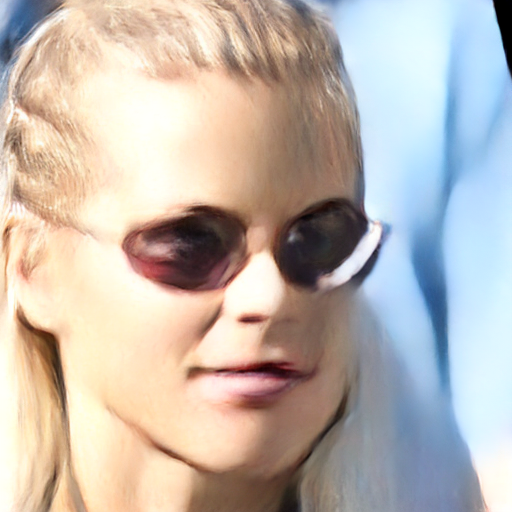} &
    \includegraphics[width=0.2\textwidth]{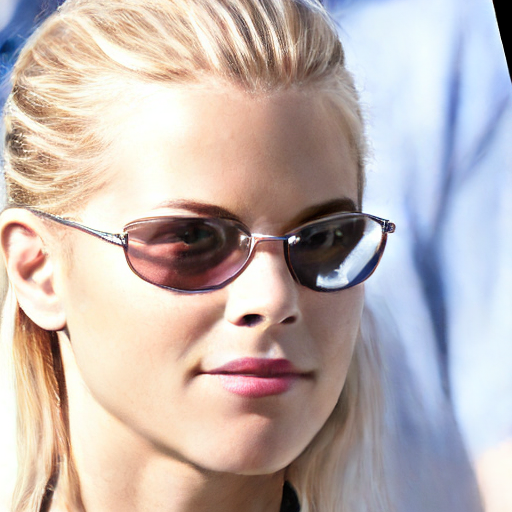} 
    \\
    LFW \imgnote  &
    OSDFace~\cite{osdface}  &
    MinMax~\cite{jacob2017quantizationtrainingneuralnetworks}  & 
    Q-Diffusion~\cite{qdiffusion}  &
    EfficientDM~\cite{He2023EfficientDM}  &
    PassionSR~\cite{zhu2024passionsr}  &
    SVDQuant~\cite{li2025svdquant} &
    QuantFace (ours) 
    \\
    \end{tabular}
    \end{adjustbox}
\vspace{-4mm}
\caption{Visual performance comparison on \textit{real-world datasets}. Please zoom in for a better view. Our QuantFace is under bit-width W4A3.97, and other methods are under W4A4.}
\label{fig:vis-realworld}
\vspace{-6mm}
\end{figure*}
\subsection{Main Results}
\vspace{-2mm}
\paragraph{Quantitative Results.}The results of the synthetic dataset are in Tab.~\ref{tab:synthetic} and the results of real-world datasets are in Tab.~\ref{table:realworld}. These results indicate that QuantFace outperforms existing post-training quantization methods across most evaluation metrics under W4A4, W4A6, and W6A6 precision settings. Notably, at W4A4 precision, our approach surpasses the state-of-the-art method on every metric.
Reference-based metrics such as LPIPS and DISTS indicate that QuantFace achieves smaller perceptual deviations from the ground-truth images, while lower LMD and Deg. further show that QuantFace restores the facial structure faithfully.
\vspace{-5.8mm}
\paragraph{Quality Results.}We show the visual comparison results of the synthetic datasets in Fig.~\ref{fig:vis-celeba} and real-world datasets in Fig.~\ref{fig:vis-realworld}. Our method reconstructs a wide range of facial details effectively, such as eyes, hair, and other elements that are prone to distortion after quantization. Moreover, even under 4-bit precision, our method still generates high-fidelity face images compared with others.
\begin{table}
\centering
\resizebox{\linewidth}{!}{
\begin{tabular}{c|ccccc}
\toprule
\rowcolor{cvprblue!30}\textbf{Ablation} &CLIP-IQA$\uparrow$  & LPIPS$\downarrow$ & Deg.$\downarrow$ & LMD$\downarrow$& \shortstack[c]{FID\\(FFHQ)}$\downarrow$\\
\midrule
SVDQuant &0.5192	&	0.4143&76.3504&10.387	&84.91\\

+ GAN Loss& 0.5374 & 0.4249 &76.8697 & 	8.852&	80.71\\

+ QD-LoRA&0.6321	&0.4060	&77.3236 &9.344	&58.88 \\

+ Rotation &\B{0.6916}	&	\B{0.3709}&\B{69.2888}	&\B{6.195}	&\B{58.34}\\

+ Mixed-Precision 	&\R{0.6984} &	\R{0.3518}&\R{65.6640}	&\R{5.927}	&\R{53.18}\\
\hline
\bottomrule
\end{tabular}
}
\vspace{-3mm}
\caption{\textbf{Ablation studies on different components starting from SVDQuant-W4A4.} Experiments are conducted on the \textit{synthetic dataset} CelebA-Test~\cite{CelebA-Test}.}
\label{tab:ablation}
\vspace{-6mm}
\end{table}

\subsection{Ablation Studies}
\vspace{-2mm}
\paragraph{Rotation-Scaling Channel Balancing.}
As illustrated in Fig.~\ref{fig:four_images}, solely applying scaling leads to uneven data distribution across channels, making it challenging to preserve the original data distribution after quantization. On the other hand, using the RHT alone fails to eliminate extreme outliers in the activations. By combining both techniques, we can effectively maintain the data distribution while reducing quantization errors. Consequently, we are able to enhance the final face restoration performance.
\vspace{-5mm}
\paragraph{Quantization-Distillation Low-Rank Adaptation.}
The results presented in the Tab.~\ref{tab:ablation} demonstrate that QD-LoRA effectively maintains a balance between quantization and distillation. QD-LoRA neither struggles to align with the FP model nor overfits to the limited calibration dataset. Therefore, models can generate excellent human face images after quantization. Results in Tab.~\ref{tab:ablation} also reveal that QD-LoRA surpasses direct optimization on the SVD low-rank branch.
\vspace{-5.5mm}
\paragraph{Adaptive Bit-width Allocation.}
Various results shown in Tab.~\ref {tab:synthetic} and Tab.~\ref{tab:ablation} demonstrate that our mixed-precision configuration enhances reference-based metrics on the synthetic dataset. The results indicate that the generated face images align closely with real-world face images. Visually, our configuration significantly improves the detail of the generated facial features, including eyes and teeth.
\vspace{1mm}
\section{Conclusion}
We introduce QuantFace, an effective quantization framework for one‑step diffusion face restoration models. QuantFace not only reduces quantization error but also facilitates the distillation from the full-precision model to the quantized model. Moreover, QuantFace promotes reasonable resource allocation. Extensive experiments demonstrate that QuantFace outperforms existing SOTA methods across a variety of metrics. By maintaining a balance between computational overhead and restoration quality, QuantFace makes the efficient deployment of advanced face restoration models possible.
{
    \small
    \bibliographystyle{ieeenat_fullname}
    \bibliography{main}
}


\end{document}